\documentclass[10pt,journal,compsoc]{IEEEtran}

\ifCLASSOPTIONcompsoc
  \usepackage[nocompress]{cite}
\else
  \usepackage{cite}
\fi

\ifCLASSINFOpdf
\else
\fi
\usepackage{ragged2e}
\usepackage[utf8]{inputenc} 
\usepackage[T1]{fontenc}    
\usepackage{url}            
\usepackage{hyperref}
\usepackage{booktabs}       
\usepackage{amsfonts}       
\usepackage{nicefrac}       
\usepackage{microtype}      
\usepackage{subfig}

\usepackage{amsmath}
\usepackage{algorithm}
\usepackage[noend]{algpseudocode}
\usepackage{ amssymb, tikz, pgfplots}

\usepackage{todo}

\begin{document}
\title{ChOracle: A Unified Statistical Framework for Churn Prediction}

\author{Ali~Khodadadi,~
Seyed~Abbas~Hosseini,~
Ehsan~Pajouheshgar,~
\\
Farnam~Mansouri,~
and~Hamid~R.~Rabiee~\IEEEmembership{Senior~Member,~IEEE}
\IEEEcompsocitemizethanks{
\IEEEcompsocthanksitem A. Khodadadi, S.A. Hosseini, E. Pajouheshgar, F. Mansouri, H.R. Rabiee are with the Department of Computer Engineering, Sharif University of Technology, Tehran, Iran Email: \{khodadadi, a\_hosseini, pajouheshgar, fmansouri\}@ce.sharif.edu, rabiee@sharif.edu.}
}

\markboth{IEEE TRANSACTIONS ON KNOWLEDGE AND DATA ENGINEERING,~Vol.~X, No.~X, Z~201X}%
{Hosseini \MakeLowercase{\textit{et al.}}: Recurrent Poisson Factorization for Temporal Recommendation}

\IEEEtitleabstractindextext{%
\begin{abstract}\justifying{
User churn is an important issue in online services that threatens the health and profitability of services. Most of the previous works on churn prediction convert the problem into a binary classification task where the users are labeled as churned and non-churned. More recently, some works have tried to convert the user churn prediction problem into the prediction of user return time. In this approach which is more realistic in real world online services, at each time-step the model predicts the user return time instead of predicting a churn label. However, the previous works in this category suffer from lack of generality and require high computational complexity. In this paper, we introduce \emph{ChOracle}, an oracle that predicts the user churn by modeling the user return times to service by utilizing a combination of Temporal Point Processes and Recurrent Neural Networks. Moreover, we incorporate latent variables into the proposed recurrent neural network to model the latent user loyalty to the system. We also develop an efficient approximate variational algorithm for learning parameters of the proposed RNN by using back propagation through time. Finally, we demonstrate the superior performance of ChOracle on a wide variety of real world datasets.

}
\end{abstract}
\begin{IEEEkeywords}
Churn Prediction, User Modeling, Marked Temporal Point Processes, Recurrent Neural Network
\end{IEEEkeywords}}

\maketitle

\IEEEdisplaynontitleabstractindextext

\IEEEpeerreviewmaketitle

%
\IEEEraisesectionheading{\section{Introduction}\label{sec:introduction}}
\IEEEPARstart{U}{sers} are the main part of any service both in online and offline worlds, and hence user acquisition and retention is of utmost importance for service providers. Recent studies show that retaining existing users is considerably less expensive than acquiring new ones, and existing users are more profitable than the newcomers \cite{ali2014dynamic, karnstedt2010churn,buckinx2005customer}. Therefore, it is common to pay more attention to user retention than user acquisition, specially in online services. Considering the growing rate of online services, \emph{user churn} which broadly refers to the loss of customers is an important issue. User churn is more challenging in online services due to the factors such as low switching costs, large number of competitors, and the free nature of many services. Therefore, many research efforts have been directed toward predicting user churn, in recent years. Once the potential churners identified, the customer relationship management (CRM) systems can target them with appropriate incentives such as tailored promotions \cite{huang2012customer} or gamification methods \cite{Khodadadi:2018:CUM} to sustain their interest in their current services.

Churn prediction is studied in different domains such as telecommunication industry \cite{yan2004predicting,lu2014customer,khan2015behavioral,huang2012customer}, banking \cite{xie2009customer}, P2P networks \cite{herrera2007modeling,Stutzbach2006Understanding}, online gaming \cite{kawale2009churn,runge2014churn}, community based question answering (CQA) services\cite{dror2012churn,pudipeddi2014user} and other online services \cite{karnstedt2010churn,lang2013social,Karumur2016Early}. There exist different definitions for churn in the literature, corresponding to various service domains. The definitions for churn can be divided into three types: \emph{Active} which is the case for subscription based services, and the churn happens when the contract is terminated and user leaves the service. \emph{Hidden} which is the case for free services and happens when the user does not use the service for a significant amount of time. And, \emph{Partial} which is the case when user do not use all the available features of the service, and instead uses those features in other services. For example, a drop in the level of user activity can be a sign of partial churn.

\begin{figure}[t] 
  \includegraphics[width=3.5in]{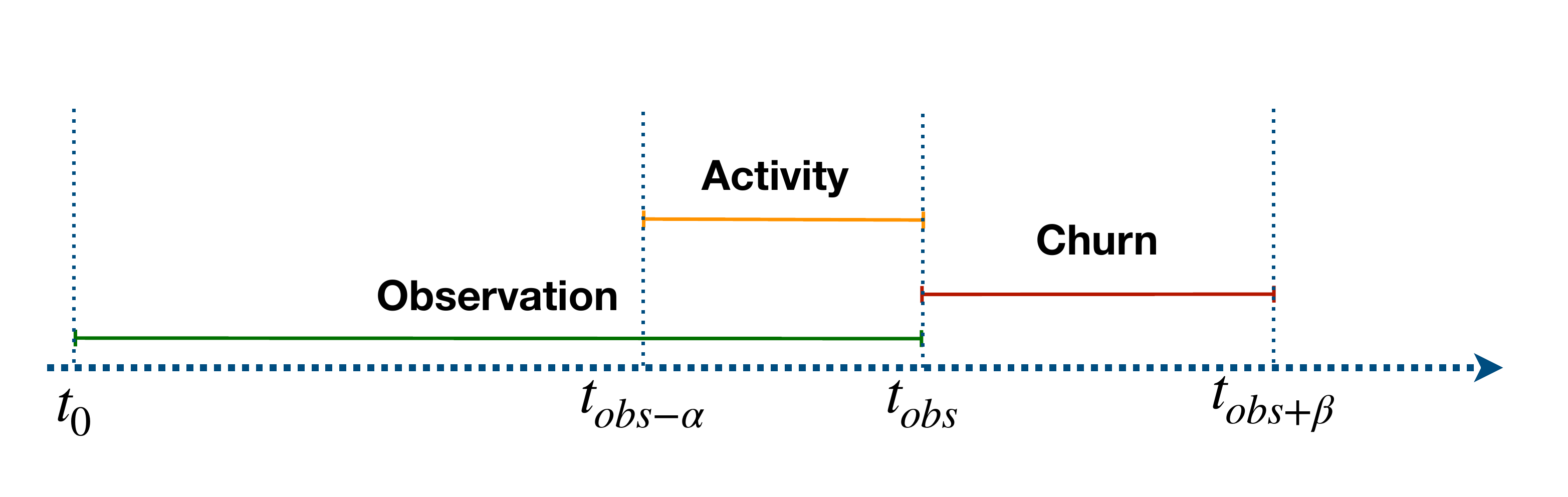}
  \caption{Churn windows in the literature. The data is gathered for the users on an \textbf{Observation} window. The model is trained based on the data for those who are active in an \textbf{Activity} window. The churn label is assigned to the users based on their activity in a \textbf{Churn} window after observation window. The length of these windows is an important parameter in churn prediction.}\label{fig:churn_windows}
\end{figure}

Most of the previous works in the churn prediction use the active and hidden definitions of churn, and convert the churn prediction problem into a binary classification problem. They usually use an \emph{observation window} in which they observe the user activity, and a \emph{churn window} in which they predict the user churn. If the user has no activity in the churn window, he is labeled as churned, otherwise it is labeled as non-churned. Fig. \ref{fig:churn_windows} illustrates the different windows used in this type of churn prediction methods. The main effort in these methods is set on feature engineering and they aggregate the user history in the \emph{observation window} into a single point or a sequence of points with a churn label, then this data is used to train a common classification method. Even though this method simplifies the application of different classical learning models on the datasets, there is many drawbacks in this approach which makes it inappropriate for many of the existing social media services. 
First, aggregating all temporal user history in a single point exposes the loss of some useful information that can be used to improve the prediction performance. Second, the choice of the thresholds for the observation and churn windows and the features selected for churn prediction are application dependent which implies that current methods could not be generalized to other domains. Moreover, the current classification based approaches can not efficiently capture the partial definition of churn, which is one of the main challenges of current online services. Moreover, in free online services, users may use different competitors simultaneously, and an decrease in usage may be a preliminary indicator of churn. In addition, the churn definition may differ from a user to another. While a level of activity for an active user may be a sign of churn, it may mean nothing for a less active user. Finally, the current approaches can not distinguish between those who churned at the beginning of churn window and the ones that departed later, while for churn prediction it is important to identify the churning time of a user just in time for making the appropriate plans to avoid churns in a timely manner.

%

Recently, some methods have approached the churn prediction problem in a different manner. They try to predict the exact returning time or accordingly exact absence gap of each user to the service for each session in order to predict if the user will leave the system or not \cite{kapoor2014hazard}. This approach do not suffer many shortcomings of the classification based methods and it utilizes the history of interactions of user with the service for churn prediction. In addition, the requirement to define observation and churn windows is mitigated. Moreover, by predicting the next interaction time of user with the service and observing deviations from the expected behavior, the service providers can detect the user churn at the early stages and also capture the partial definitions of churn. 
The authors in \cite{kapoor2014hazard} proposed a hazard model from survival analysis to predict user return times to the service. For the first time, they converted the classification approach in this literature into a regression problem, and used the survival analysis to solve the problem. Though innovative, their proposed method suffers from some weaknesses. First, their proposed method needs many feature engineering efforts, and is well suited for music streaming services, and is not applicable to other services. Second, for each session they aggregated the user history data into a single point with some features and lose many temporal information about the exact timing of previous events.
The authors in \cite{NeuralSurvival} also proposed the neural survival recommender. Their proposed method, maps the gap between the user sessions into hourly beans, then it learns a very large vector of these hourly intensities by using recurrent neural networks. Though their focus is on recommender systems but their approach can be utilized for returning time predictions. This method also suffers from some drawbacks. It needs to learn very high dimensional vectors for each session that may impose curse of dimensionality when the number of events per user are not very high. Moreover, it quantizes the temporal gaps between sessions into hourly beans that may result in loss of information.

In this paper, we introduce \emph{ChOracle}, an oracle to predict the churn which is able to resolve the aforementioned challenges in the previous works. The proposed method models the future absence gaps of users from the service by utilizing the timing and durations of previous sessions. It exploits the strangeness of temporal point processes and recurrent neural networks in an innovative session based manner to model the return times, session to the service and session durations. Since it only uses user return times and session duration information, it can be considered as a general purpose method that can be used in a broad range of online services. The overall contributions of this paper are:
\begin{itemize}
	\item We analyze the users interactions with the service in granularity of \emph{sessions}, and propose to use user return times and session durations to predict user churn.
	\item We propose a special Marked Temporal Point Process to model user return times to the service or accordingly their absence gaps and predict session durations. 
	\item We utilize RNNs to code the history of user interactions with the service and define the intensity function of proposed temporal point process. Thus the proposed method is not bounded to parametric forms of intensity function, and is able to define and learn general intensity functions. We also incorporate latent random variables into the proposed RNN to add more flexibility to the model. This increases the expressive power of the proposed method and make it capable of handling highly structured data. 
	\item We introduce a variational lower bound for the complete log likelihood as the objective function, and propose an approach to maximize it through back propagation through time (BPTT).
	\item We conduct several experiments on real datasets to demonstrate the performance of proposed method. To this end, we utilize several datasets from different domains.\footnote{The codes and data are available at \href{https://github.com/alikhodadadi/ChOracle}{https://github.com/alikhodadai/ChOracle.}}
\end{itemize}

{\bf Related Work}.
The works most closely related to the proposed method can be divide into two groups: Churn prediction methods and Temporal point process based methods. These are two independent lines of research and to the best of our knowledge we are the first to systematically combine these two and propose a unified statistical framework for churn prediction.

Churn prediction is studied in a wide range of domains, including telecommunications \cite{yan2004predicting,lu2014customer,khan2015behavioral,huang2012customer}, online games\cite{kawale2009churn,runge2014churn}, P2P networks \cite{herrera2007modeling,Stutzbach2006Understanding}, and community based question answering (CQA) services \cite{dror2012churn,pudipeddi2014user}. The current literature mainly uses the hidden definition of churn and convert the problem into a binary classification problem. The most common topic that is addressed by the previous work is focused on evaluating different predictive models that are trained on a domain-specific dataset. In this context, the main effort is set on engineering appropriate features for the task and different machine learning methods are evaluated such as decision trees \cite{pudipeddi2014user,wei2002turning}, logistic regression \cite{buckinx2005customer,ali2014dynamic}, neural networks \cite{tsai2009customer,runge2014churn}, random forests \cite{dror2012churn,xie2009customer}, and support vector machines \cite{coussement2008churn}. Hence, they are very application dependent and suffer from the lack of generalization. 
Most of the previous works use the intrinsic features of users and service, and aggregate the features over the entire observation period and pay little attention to the temporal aspects of user behavior. While this approach simplifies the problem, but it results in loss of some important information about the temporal aspects of user behavior that can be utilized in churn prediction.
The authors in \cite{martins2017predicting} used some temporal aspects of data and considered the problem as a sequence classification problem, and proposed a recurrent neural network to classify the sequences. While using temporal information, their approach suffers from many drawbacks.
The authors in \cite{karnstedt2010churn} used the partial definition of churn and tried to use social network features on churn prediction but they did not paid any attention to the temporal aspects of user behavior.
To conclude, the previous works on classification based methods for churn prediction, mainly focused on the hidden definition of churn and little attention is paid to partial definition of churn while this is the case for current social media services. They used popular classification based methods for classification, and temporal information about the user interactions with the service is neglected while it is important for better prediction of the user churn.

On the other hand, survival analysis and point process based methods have attracted a lot of interest in different mining problems such as modeling marked temporal events \cite{HNP3,Coevolve,farajtabar2014shaping,RMTPP}, user behavior modeling \cite{Khodadadi:2018:CUM}, modeling the diffusion of products \cite{C4} and recommendation systems~\cite{CoevolutionaryRecommenderSystem,hosseini2017RPF,NeuralSurvival}. Most of these works try to model user return time to enhance another primary task such as recommendation, and are not focussed on modeling the return time itself. Besides, they mainly model the user return times to the items, while in the churn prediction we are interested in predicting the user return time to the service not individual items. 
They also try to model the mark of events as a countable finite variable, while we propose to model the session duration as an infinite continuous random variable. 
The works in \cite{kapoor2014hazard} and \cite{NeuralSurvival} are the ones that are  somehow related to our work. To the best of our knowledge, Kapoor et al. in \cite{kapoor2014hazard} for the first time approached the churn prediction problem in a return time prediction manner and tried to model the user return time to the service. They used the Cox proportional hazard function to model user return times. While simple, their proposed method suffers from many shortcomings. To name a few, they aggregated the user history in a single point which results in loss of many temporal information. Furthermore, their approach needs many feature engineering efforts which makes it applicable to a specific service and  endangers the generality of their approach. The authors in \cite{NeuralSurvival} also tried to model user return times to the service. But, their primary goal is the recommendation task. They proposed to model user return times to the service not the items, and proposed a method for next-basket recommendation. They learn a large vector of fixed intensities binned for every hour, and hence their method suffers from high computational complexity. To conclude, the literature on point processes mainly do not focussed on the return time prediction as the main goal, but as a covariate for other primary tasks. They mainly modeled the return times to the items not the whole service and the proposed point processes have single purpose parametric forms which suffers from lack of generality. On the other hand, we propose a general purpose model, that only uses user return times and session durations information to predict user return times to the service. However, our proposed method concentrates on service level and  do not need any feature engineering efforts and do not suffer from high computational complexity.
\section{Preliminaries}\label{sec:Preliminaries}
\subsection{Temporal Point Processes}
A temporal point process is a powerful mathematical tool for modeling random events over time. More formally, a temporal point process is a stochastic process whose realizations consists of a list of time-stamped events $\{t_1, t_2, \ldots, t_n\}$ with $t_i\in \mathbb{R}^+$. Different types of activities over a service, can be considered as events generated by a point process. In our case, the sessions of a user in the service can be considered as the events generated by the process.

The length of the time interval between successive events is referred to as the inter-event duration. A temporal point process can be completely specified by distribution of its inter-event durations \cite{PointProcessesTheory}. Let $\mathcal{H}_t$ denote the history of events up to time $t$, then by applying the chain rule we have:	
\begin{align}\label{eqn:CompactLikelihood}
f(t_1, \ldots, t_n) = \prod_{i=1}^n f(t_i | t_1, \ldots, t_{i-1}) = \prod_{i=1}^n f(t_i|\mathcal{H}_{t_i})  
\end{align}
Therefore, to specify a point process, it suffices to define  $f^*(t)=f(t|\mathcal{H}_t )$, which is the conditional density function of an event occurring at time $t$ given the history of events.

A temporal point process can also be defined in terms of counting process $N(t)$ which denotes the number of events up to time $t$. The increment of the process, $dN (t)$, in an infinitesimal window $[t, t + dt)$, is parametrized by the conditional intensity function $\lambda^*(t)$.  The function $\lambda^*(t)$ is formally defined as the expected rate of events occurring at time $t$ given the history of events, that is:
\begin{align}
	\lambda^*(t)dt = \mathbb{E}[dN(t)|H_t]
\end{align}
There is a bijection between the conditional intensity function (intensity for short) and the conditional density function:
\begin{align} \label{eqn:lambda}
\lambda^*(t) = \frac{f^*(t)}{1-F^*(t)}
\end{align}
where $F^*(t)$ is the Cumulative Distribution Function (CDF) of $f^*(t)$. Using the definition of $\lambda^*(t)$ in Eq.\ref{eqn:lambda}, the likelihood of a list of events ($t_1, \ldots, t_n$) which is observed during  a time window $[0,T)$, can be defined as: 
\begin{align} \label{eqn:PPLikelihood}
\mathcal{L} = \prod_{i=1}^{n} \lambda^*(t_i) \exp \left(-\int_0^T\lambda^*(s) ds \right)
\end{align}
where $n$ is the number of observed events and $T$ is the duration of observation. 
Intuitively, $\lambda^*(t)$ is the probability of an event occurring in time interval $[t,t+dt)$ given the history of events up to $t$, and it is a more intuitive way to characterize a temporal point process by its intensity function \cite{Aalen2008}. For example, a temporal Poisson process can be characterized as a special case of a temporal point process with a history-independent intensity function which is constant over time, i.e. $\lambda^*(t) = \lambda$ \cite{kingman1993poisson}. Users' actions usually exhibit complex longitudinal dependencies such as self-excitation, where a user tends to repeat what he has done recently. Such behavioral patterns can not  be characterized by a homogeneous Poisson process, and hence more advanced temporal point processes are needed. Hawkes process is a temporal point process with a particular intensity function which is able to capture the self-excitation property. The intensity function of a Hawkes process is given by:
\begin{align}\label{eqn:hawkes}
\lambda^*(t) = \mu+\alpha g_{\omega}(t)\star dN(t)=\mu + \alpha \sum_{t_i<t} g_{\omega}(t-t_i)
\end{align}
where $\mu$ is a constant base intensity, $\alpha$ is a weighting parameter which controls the impact of previous events on the current intensity, $g_{\omega}(t)$ is a kernel which defines the temporal impact of events on the future intensity, and $\star$ is the convolution operator. In the case that $g_\omega(t)$ is a decreasing function, Hawkes process produces clustered point patterns over time and hence is able to model the self-excitation property of users events. The right hand side of Eq. \ref{eqn:hawkes} comes from the fact that the number of events occurred in a small window $[t,t+dt)$ is $dN(t) = \sum_{t_i\in \mathcal{H}_t}\delta(t-t_i)$, where $\delta(t)$ is a Dirac delta function. 

Each event can also be associated with some auxiliary information known as the mark of an event. For example, the song user played in a music streaming service, the item user bought in an online retailer service, or the question user answered in a CQA website can be considered as the marks of events. In our case, the user sessions can be considered as the events and the duration of sessions can be considered as the mark of events. A marked temporal point process is a point process for modeling such events. If $k$ denotes the mark of the events, then the intensity of the marked temporal point process is given by:
\begin{align}\label{eqn:markedLambda}
\lambda^*(t,k) = \lambda^*(t) f^*(k|t) 
\end{align}
where $\lambda^*(t)$ denotes the temporal intensity function, and $f^*(k|t)$ is the conditional probability density function of observing an event with mark $k$ at time $t$. Therefore, in order to determine a temporal point process, we need a temporal intensity which shows the rate of occurring event given the history and a conditional probability density function over marks. In the marked case, the likelihood of a list of events \{$(t_1,k_1), \ldots, (t_n,k_n)$\} which is observed during  a time window $[0,T)$, can be defined as: 
\begin{align} \label{eqn:PPLikelihood}
\mathcal{L} = \left(\prod_{i=1}^{n} \lambda^*(t_i,k_i)\right) \exp \left(-\int_0^T\lambda^*(s) ds \right)
\end{align}

\subsection{Recurrent Neural Networks}
Recurrent neural networks (RNNs) are a class of neural networks suited for dealing with sequential data. In traditional neural networks we assume that all inputs are independent of each other which do not holds in many real world situations. For instance, if we want to predict the next word in a sequence, it is highly correlated with the words that came before it, or if we want to predict the next location a user will check in, it is also correlated with the locations he has previously checked in.
RNNs are feed-forward neural networks that allow connections between hidden units. More specifically, additional edges are added to the network so that the hidden state of the current time step is fed as the input to the network in the next time step. As a result the same structure is repeated at adjacent time steps through time, and hence they are called recurrent neural networks. This small modification results in an important property for RNNs: The hidden state at the current time step, depends on all the previous hidden states and hence all the previous inputs. This memory property is the main engine of RNNs that give them an excellent predictive power.  

\begin{figure}[!t]
  \includegraphics[width=3.5in]{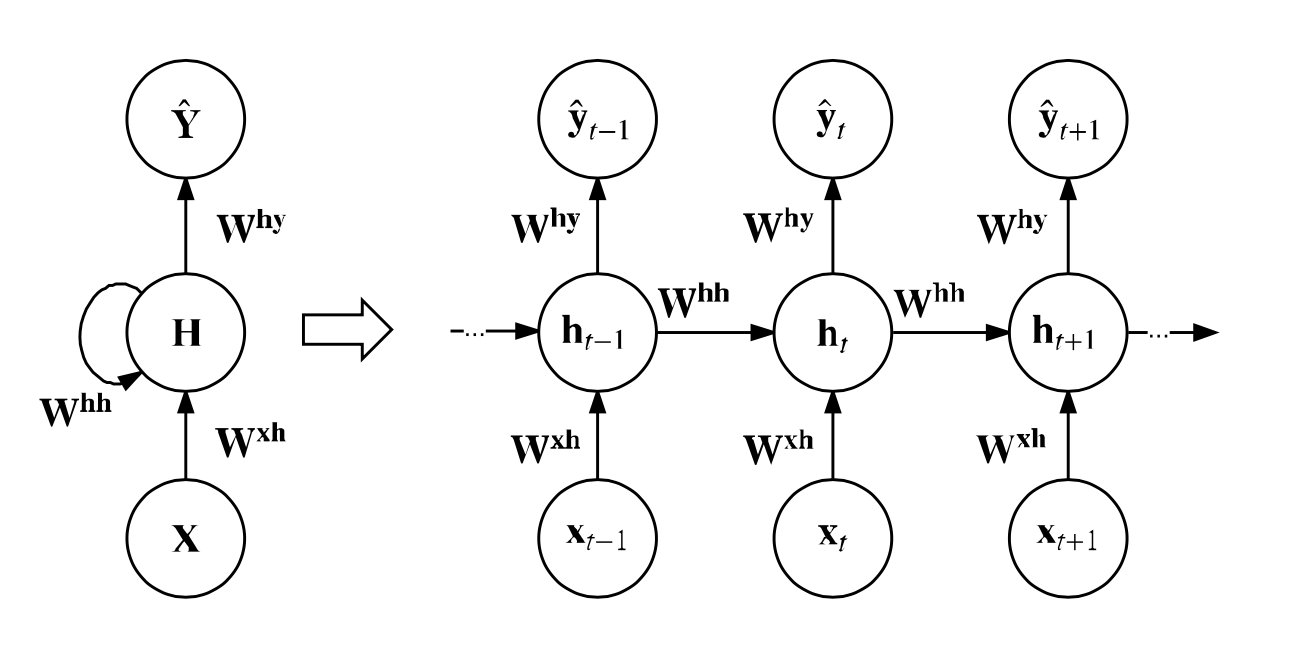}
  \caption{An RNN unrolled through the time. The same structure is repeated at adjacent time steps.}
  \label{fig:RNN-Unrolled}
\end{figure}

Let consider $h_t$ and $x_t$ as the hidden state and input at time step $t$, respectively. The hidden state depends on both the current input and the hidden state of previous time step as follows: 
\begin{align}
	h_t =  \sigma(W^{xh}x_t + W^{hh}h_{t-1} + b_h)
\end{align}
where, $W^{xh}$ and $W^{hh}$ are the weight matrices that connects the input to the current hidden state, and previous hidden state to the current hidden state, respectively. $b_h$ is the bias vector, and $\sigma(.)$ is the activation function. The Sigmoid, hyperbolic tangent, and ReLU are examples of popular activation functions. The initial state $h_0$ only depends on the input, and each state can be calculated in a similar fashion. Fig. \ref{fig:RNN-Unrolled} illustrates the recurrent nature of and RNN that is unrolled through the time. The weight matrices are learned using training data by an approach called back propagation through time (BPTT).

In practice, RNNs has shown state of the art performance in general purpose sequence modeling tasks such as sequence-to-sequence translation \cite{sutskever2014sequence}, handwriting recognition \cite{graves2009novel}, image captioning \cite{vinyals2015show}, and discrete time-series prediction \cite{chandra2012cooperative}.
\begin{figure*}[!t]
\centering
	\includegraphics[width=6in]{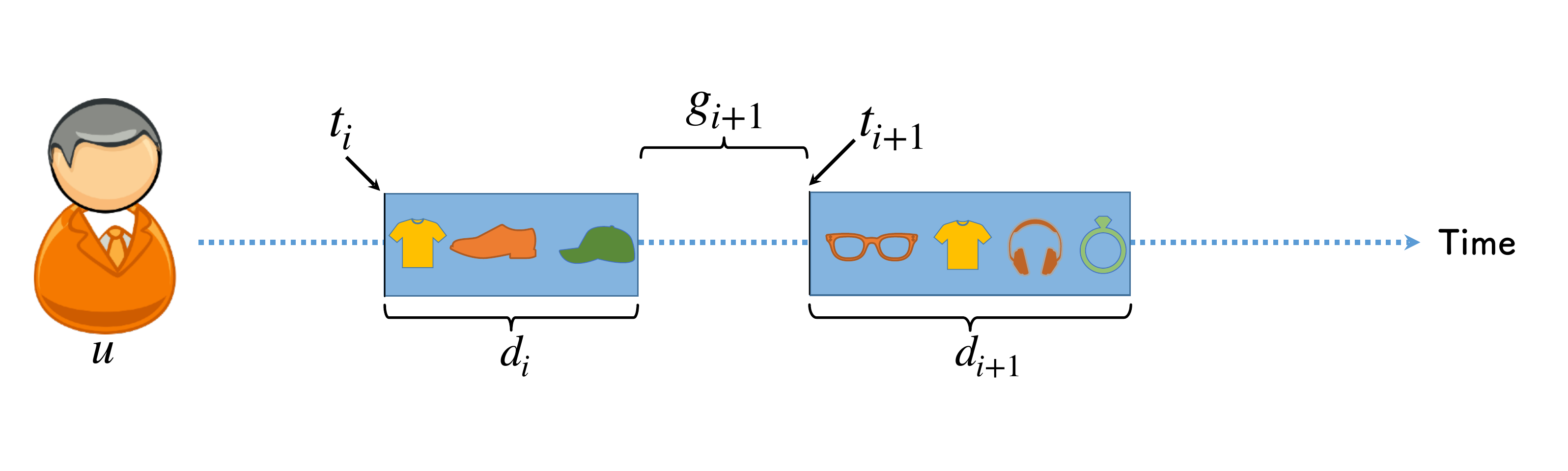}
	\vspace{-5mm}
	  \caption{ Illustration of user session data. Each session $i+1$ has a starting point $t_{i+1}$, a previous absence gap $g_{i+1} = t_{i+1}-t_{i}$, which is the gap between consecutive sessions, a duration $d_{i+1}$ in terms of time the user spent on the service or the number of items user consumed. In the first case the $d$ is a continuous variable, while in the second it is an infinite discrete random variable. }\label{fig:Sessions}
\end{figure*}
\section{ ChOracle: A churn prediction framework based on RNNs}
\label{sec:proposed}
As we mentioned in the introduction section, a recent approach in churn prediction is to predict the returning time of the user to the service. Temporal Point Process (TPP) is a strong mathematical framework for modeling the underlying patterns governing the temporal data. The major limitation of existing studies that use TPPs to model temporal data is that they often draw parametric assumptions about the conditional intensity function. Each parametric form determines different temporal characteristics for the point process and tries to correctly decide which form to use. This is a hard task that needs sufficient domain knowledge. The authors in \cite{RMTPP} proposed a unified model that do not need parametric assumptions about the intensity function. To this end, they combined the temporal point processes and recurrent neural networks by letting the RNN to determine the value of intensity function of temporal point process at each time-step. Though innovative, their approach suffers a limitation caused by RNNs. The hidden state in an RNN is a deterministic function of input and the previous state, and the only source of variability is the conditional output probability function which can limit the expressiveness of the model for highly structured data where some latent random variables govern the temporal dynamics of data\cite{VRNN}. 

To tackle this challenge, in this section, we propose a mathematical framework to define the conditional intensity function of a temporal point process using recurrent neural networks. We incorporate latent random variables into the recurrent neural network to model the variability observed in the data. The latent variable can be interpreted as the latent loyalty of users to the service that increases the expressive power of proposed method. Hence, the proposed method, is able to better predict the returning times of users to the system and user churn. In the followings, we first specify some notations and then discuss the proposed framework.

\subsection{Notations and Conventions}
We cast the churn prediction problem as a return time prediction problem. Moreover, in order to predict user returns to the service, it is better to model user sessions in a system level instead of detailed user events. Besides, we argue that knowing how long the user stayed at the system in each session in addition to the true return times, can improve the predictions and help to better model future return times and the user churn.
Let $\mathcal{H}(T) = \{\mathcal{S}^u(T)\}_{u=1}^{U}$ denote the collection of all sequences of user sessions in the system up to time $T$, where $U$ is the total number of users and $\mathcal{S}^u(T)$ is the sequence of sessions of user $u$ till time $T$. That means $\mathcal{S}^u(t)=\{s^u_i\}_{i=1}^{n^u(t)}$, where, $s^u_i$ is the $i$th session of user $u$ in the system and is denoted as $s^u_i =(t^u_i, d^u_i)$, where $t^u_i$ is the time that the session $i$ is started, and  $d^u_i$ is the duration of session. $d^u_i$ can either be the duration of the session in terms of time or the number of actions the user taen in the system. Fig. \ref{fig:Sessions} provides some intuitions about the definitions of session duration, absence gap and session start times in our model.
%
\begin{figure*}[!t]
\includegraphics[scale=.5]{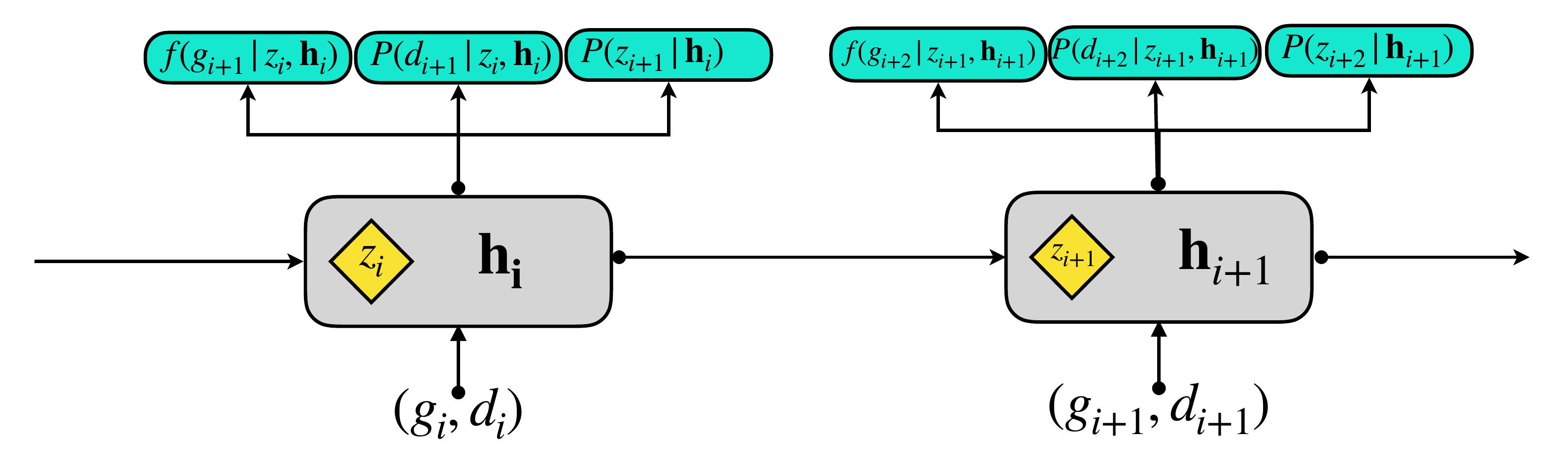}
\vspace{-3mm}
\caption{Illustration of proposed RNN based temporal point process. The inputs are the previous absence gap ($g_i$) and the session duration ($d_i$), and the outputs are the probability distributions of next absence gap, next session duration and the next latent variable. }\label{fig:proposed_rnn}
\end{figure*}
\subsection{Model Formulation}
The key idea is to model both absence gap and session duration in order to predict churns through return times. To this end, we propose a temporal point process to jointly model the next return time and session duration given all previous sessions of a user. To be able to capture general dependencies between the current and previous sessions, we let an RNN to jointly model the nonlinear dependency of current return time and session duration to the past sessions. Indeed, the output of RNN at each time-step will define the intensity function of temporal point process. To add more flexibility and expressiveness to the proposed model, we add latent random variables to the RNN. In the context of churn prediction, this latent variables can be interpreted as the current loyalty of user to the system. This latent variables will make the proposed method capable to deal with highly structured data. Fig. \ref{fig:proposed_rnn} presents the overall view of the proposed RNN based TPP. 

Given a sequence of events, $\mathcal{S}^u(t) =\{(t^u_i, d^u_i)\}_{i=1}^{n^u(t)}$, at time-step $i$, the tuple $(g^u_i, d^u_i)$ is fed into the RNN, where $g^u_i$ is the current gap, and is calculated as $g^u_i = t^u_i - t^u_{i-1}$, and $d^u_i$ is the duration of session $i$. We also consider a latent random variable $z^u_i$, which reflects the current loyalty of user $u$ to the system at step $i$. From now on, for simplicity we remove the superscript $u$ from inputs and variables. Using the aforementioned inputs $g_i$, $d_i$ and variable $z_i$, the update equation for the hidden state of RNN at time-step $i$ will be as follows:
\begin{align} \label{rel:hidden_state_update}
	\mathbf{h}_{i} = f_{\theta}(g_i, d_i, z_i, \mathbf{h}_{i-1})
\end{align}
where, $g_i$ and $d_i$ are the current inputs, $z_i$ is the current value of latent random variable, and $\mathbf{h}_{i-1}$ is the hidden state of RNN at the previous time step. $f_{\theta}$ can be any activation function such as Sigmoid ($\sigma(.)$) or ReLU ($\mathbf{R}(.)$). The output of RNN at  time-step $i$ is the probability density functions of $g_{i+1}$ and $d_{i+1}$. In the following, we describe how $p(g_{i+1}|z_i, \mathbf{h_i})$ and $p(d_{i+1}| z_i, \mathbf{h}_i)$ can be calculated. As we mentioned, we use temporal point processes to model the return times to the system and let RNN to define the conditional intensity function of the temporal point process at each time-step. Hence, to define $p(g_{i+1}|z_i, \mathbf{h_i})$, we should define $\lambda^*(g_{i+1}|z_i, \mathbf{h_i})$ of associated point process. We propose the following formulation for the conditional intensity function: 
\begin{align}\label{eq:lambda}
\lambda^*(g|z_i, \mathbf{h}_{i}) = \exp\left(w^z z_i+w^h \mathbf{h}_{i} + w^tg+b^t\right)
\end{align}
where, $\exp$(.) is used to meet the required condition for conditional intensity function to be always positive ($\forall t, \lambda^*(t) \geq 0$).
We can define the probability density function of next absence gap using the defined intensity function:
\begin{align}
g_{i+1}|z_i, \mathbf{h}_{i} \sim f^*(g|z_i, \mathbf{h}_{i})
\end{align}
\begin{align} \label{rel:next_event_prob}
f^*(g|z_i, \mathbf{h}_i) &= \lambda^*(g|z_i, \mathbf{h}_{i})\exp(-\int_{0}^g \lambda^*(s|z_i, \mathbf{h}_{i}) ds) \nonumber\\
\end{align}
It is worth noting that since $\mathbf{h}_i$ is a function of previous hidden states that depend on the related inputs, (relation \ref{rel:hidden_state_update}), $f^*(g|z_i, \mathbf{h}_i)$ will depend on all previous history. i.e. $f^*(g_{i+1}|z_i, \mathbf{h}_i) = f^*(g_{i+1}|z_{\leq i}, g_{\leq i}, d_{\leq i})$.
 
In the same way, we try to model the duration of next session at each time step. We consider $d_{i+1}$ as the number of events occurred during the session $i+1$. We also assume that it has a Poisson distribution with a parameter $\gamma$. Therefore, at each time-step $i$ we need to model the parameter $\gamma_i$ to model the next session duration. We propose the following formulation for $\gamma_i$:
\begin{align}
\gamma_{i} = \exp(w^{z,\gamma}z_i + w^{h,\gamma}\mathbf{h}_{i} + b^\gamma)
\end{align}
 Where, $w^{z,\gamma}, w^{h,\gamma}$ are the weight vectors that will be learned through the learning of RNN. Therefore, the probability density function of next session duration can be defined as:
\begin{align}
p(d_{i+1} = k|z_i,\mathbf{h}_{i}) = \frac{\gamma^k_{i}e^{-\gamma_{i}}}{k!}  
\end{align}

We also consider the latent variable $z_i$ to define the loyalty to the system in time step $i$ and hence we consider it in $[0,1]$ interval, where 0 means no loyalty to the system, and 1 is the highest loyalty and engagement with the system.  To this end, we consider a prior \emph{logit-normal} distribution for z:
\begin{align}
p(z_i| \mathbf{h}_{i-1})= p_{\theta_p}(z_i| \mathbf{h}_{i-1}) = \mathit{P} (\mathcal{N}(\mu_0, \sigma^2_0)), \\\nonumber
\text{where}, \quad\theta_p = \{\mu_0, \sigma^2_0\}
\end{align}
In order to generate this distribution, at each time-step of RNN we use a multilayer perceptron network (MLP) that accepts the previous hidden state as input and outputs $\mu_0$ and $\sigma_0$. The weights of this MLP is also shared between all time-steps and are learned through the learning phase of RNN.

\subsection{Parameter Learning}
Given the collection of all sequences of events and their latent random variables until time $T$, i.e. $\{\mathcal{S}^u(T), \mathcal{Z}^u(T)\}_{u=1}^U$, where $\mathcal{S}^u(T)$ is the sequence of sessions of user $u$, and $\mathcal{Z}^u(T)$ is the sequence of latent random variables of user $u$, we can define the joint log-likelihood as:
\begin{align} \label{eq:joint_complete_log_Likelihood}
\mathcal{L}\Big(\Big\{\mathcal{S}^u(T),& \mathcal{Z}^u(T)\Big\}_{u=1}^{U} \Big)= \nonumber\\
&\sum_{u} \log P \Big(\mathcal{S}^u(T)|\mathcal{Z}^u(T) \Big) + \log P\Big(\mathcal{Z}^u(T)\Big) \nonumber\\
&=\sum_{u}\sum_{i} \bigg( \log P(g^u_i| z^u_{i-1}, \mathbf{h}_{i-1}) + \nonumber\\
&\log P(d^u_i | z^u_{i-1}, \mathbf{h}_{i-1}) 
+ \log P(z^u_i |\mathbf{h}_{i-1}) \bigg) 
\end{align}
where, the first term is the probability density of event timings or absence gaps, the second term is the probability density of session durations, and the third term is the prior probability of latent variables.

\begin{figure}[!t]
\includegraphics[scale=.2]{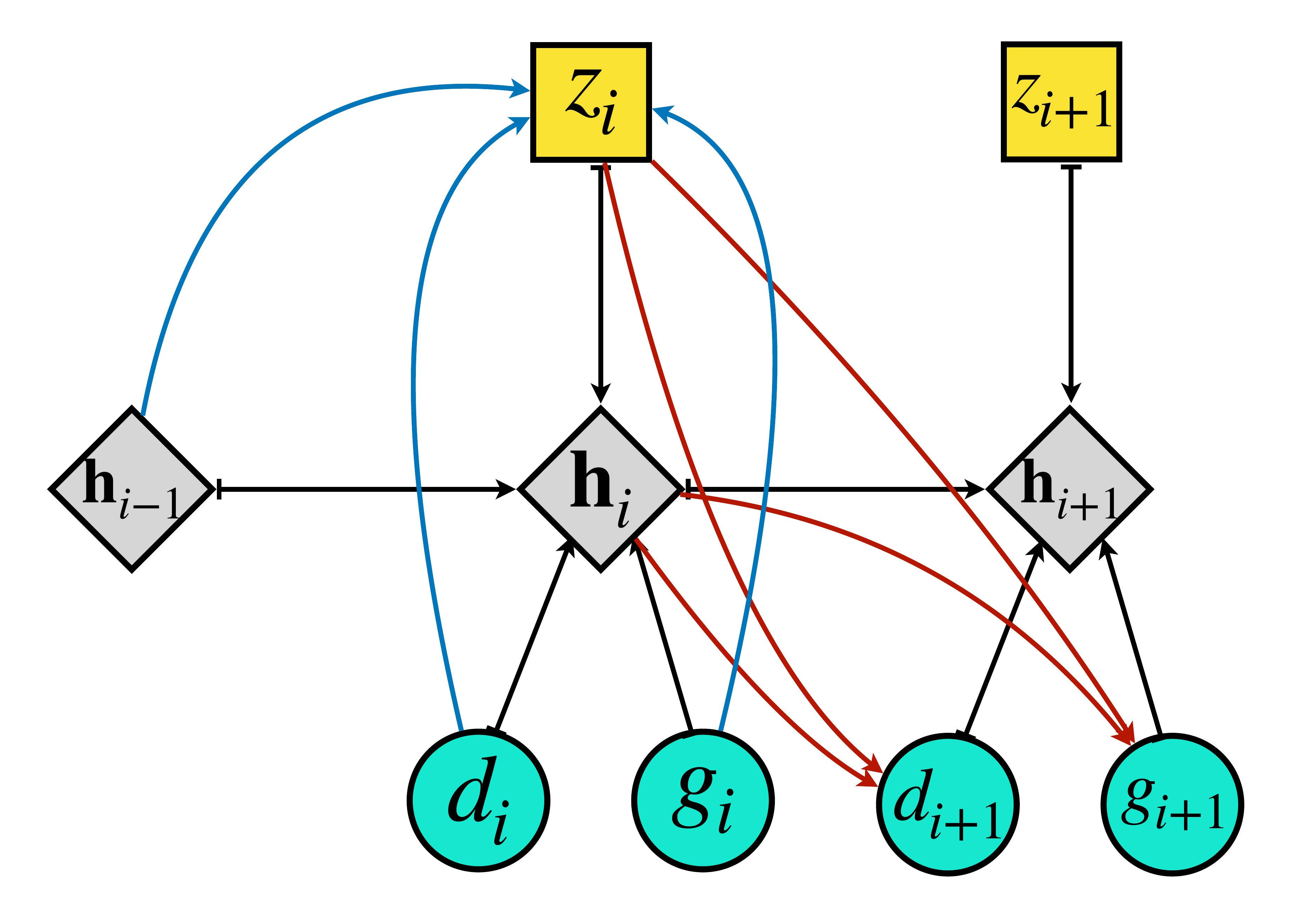}
\vspace{-3mm}
\caption{Graphical model of proposed inference method. \textbf{Black} lines show how the hidden state is inferred. \textbf{Red} lines show how the next absence gap and session duration is predicted, and \textbf{Blue} lines show how the latent variable is inferred.}\label{fig:graphical_model}
\end{figure}

Given the latent random variables, we can simply maximize this joint log-likelihood by using the Back Propagation Through Time (BPTT). However, in real data we do not have the latent variables at hand, and hence we can not directly maximize the complete joint log-likelihood in Eq. \ref{eq:joint_complete_log_Likelihood}. Utilizing the approach used in Variational Inference \cite{wainwright2008graphical} and Variational Auto-Encoder (VAE) \cite{VAE}, we try to maximize the evidence lower bound (ELBO) as the objective function. The objective function will be a time-step-wise variational lower bound as follows:
\begin{align} \label{eq:ELBO}
&\sum_{u}\Bigg\lbrack
\mathbf{E}_{q(z^u_{1:T}|g^u_{1:T},d^u_{1:T})} \bigg\lbrack\sum_{i=1}^T \bigg( \log P(g^u_i|z^u_{i-1}, \mathbf{h}_{i-1}) + \nonumber\\
&\log P(d^u_i|z^u_{i-1}, \mathbf{h}_{i-1})
-\mathbf{KL}\Big(q(z^u_{i}|g^u_{i}, d^u_{i}, \mathbf{h}_{i-1}) \parallel p(z^u_{i}|\mathbf{h}_{i-1})\Big)\bigg)\bigg\rbrack \Bigg\rbrack
\end{align}
Where $q(.)$ is the approximate posterior of latent random variables $z$. We also define the approximate posterior to be a \emph{logit-normal} distribution, and define it by modeling its parameters through \textbf{MLP}s. By maximizing this variational lower bound with respect to corresponding parameters, all the models will be learned. We also use BPTT to learn the variational lower bound. Fig. \ref{fig:graphical_model} illustrates the graphical model of the proposed model. The exact derivation of Eq.\ref{eq:ELBO} is provided in Appendix A.

%
\subsection{Churn Prediction}
It is worth mentioning that predicting the next absence gap and the next session duration is the first step in our churn prediction, and alarms can be triggered based on application needs, at the next step. For instance, in the simplest case, if the predicted values for the next session exceeds some thresholds, i.e. if $\hat{g}^u_{i+1}> \theta_g \wedge \hat{d}^u_{i+1}< \theta_d$, where $\theta_g$ and $\theta_d$ are the predefined thresholds, then the alarms can be triggered. This is equivalent to the latent definition of churn in the literature. One can also consider some more complex thresholds. For example, the thresholds can be defined based on the expected behavior of user as $\hat{g}^u_{i+1}> \mathbf{E}\lbrack g^u \rbrack  \wedge \hat{d}^u_{i+1}>\mathbf{E}\lbrack d^u \rbrack$ which is equivalent to the partial definition of churn.
Since choosing this thresholds is application dependent, we concentrate on prediction through exact values, and leave the detailed investigation of churners and churn alarms to future works. 
\begin{figure*}[!t]
\makebox[100pt][s]{\bf Tianchi}
\hspace{0.2in}%
\makebox[100pt][s]{\bf Last.fm}
\hspace{0.2in}%
\makebox[100pt][s]{\bf Foursquare}
\hspace{0.6in}%
\makebox[-50pt][s]{\bf IPTV}
\\
\centering
\subfloat[]{\includegraphics[height=1.3in,width=1.75in]{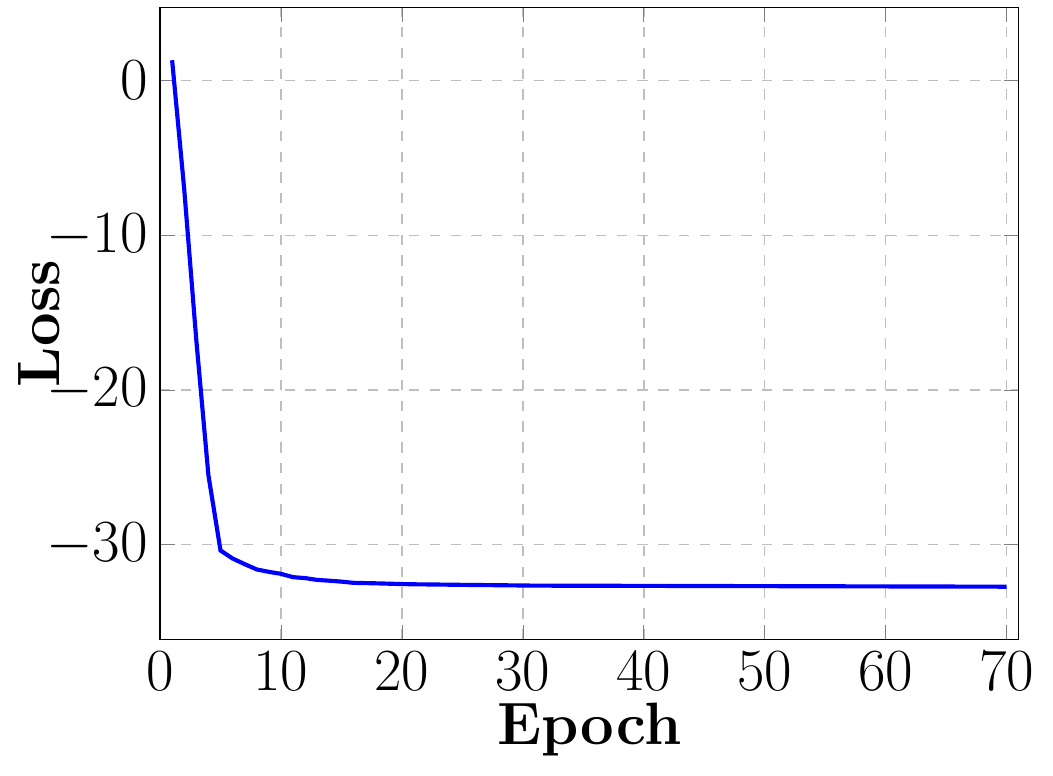}%
\label{fig:loss_vs_epoch_tianchi}}
\hspace{0.01in}%
\subfloat[]{\includegraphics[height=1.3in,width=1.75in]{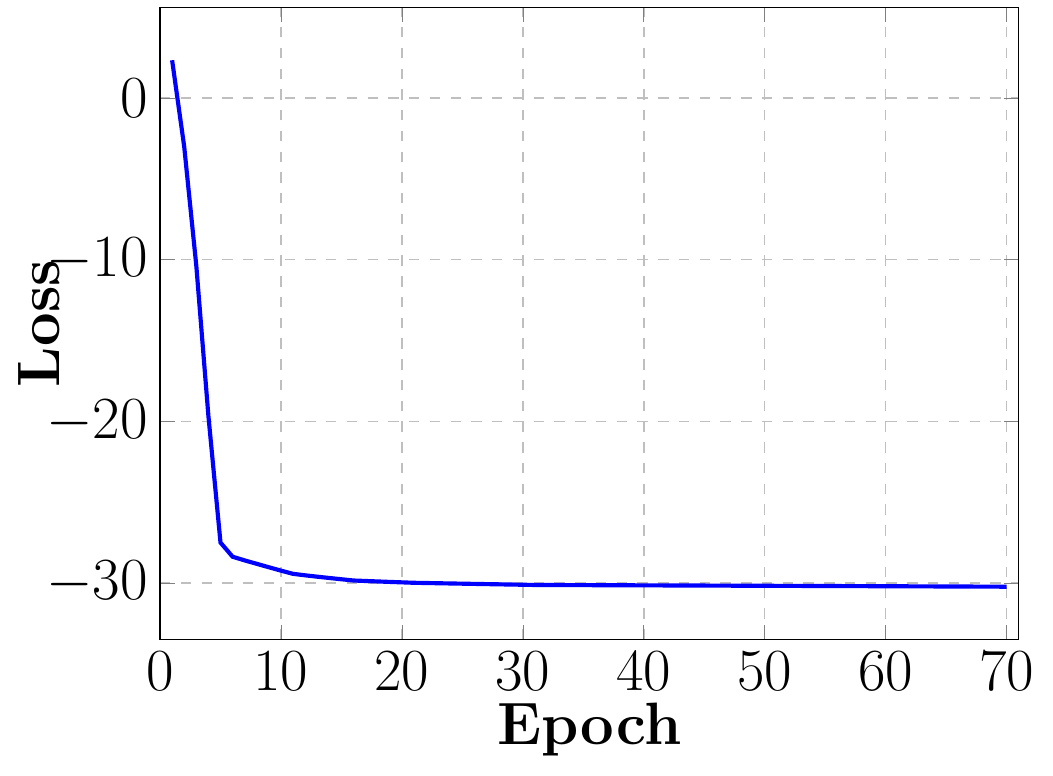}%
\label{fig:loss_vs_epoch_lastfm}}
\hspace{0.01in}%
\subfloat[loss vs sample size]{\includegraphics[height=1.3in,width=1.75in]{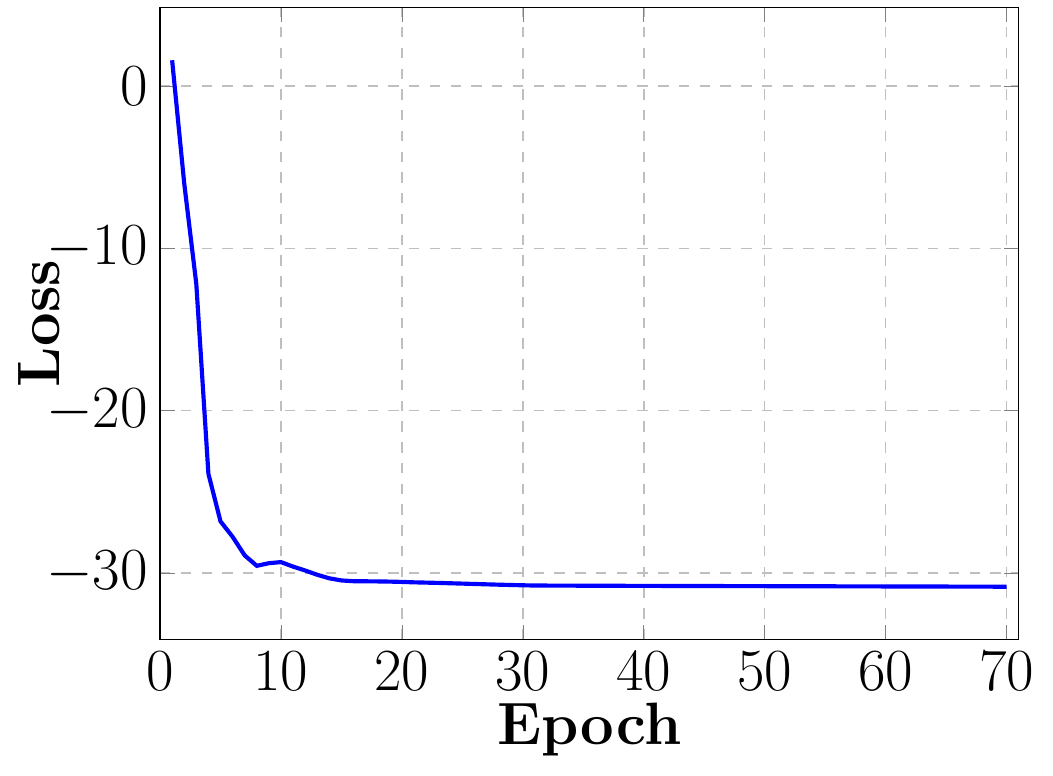}%
\label{fig:loss_vs_epoch_foursquare}}
\hspace{0.01in}%
\subfloat[]{\includegraphics[height=1.35in,width=1.75in]{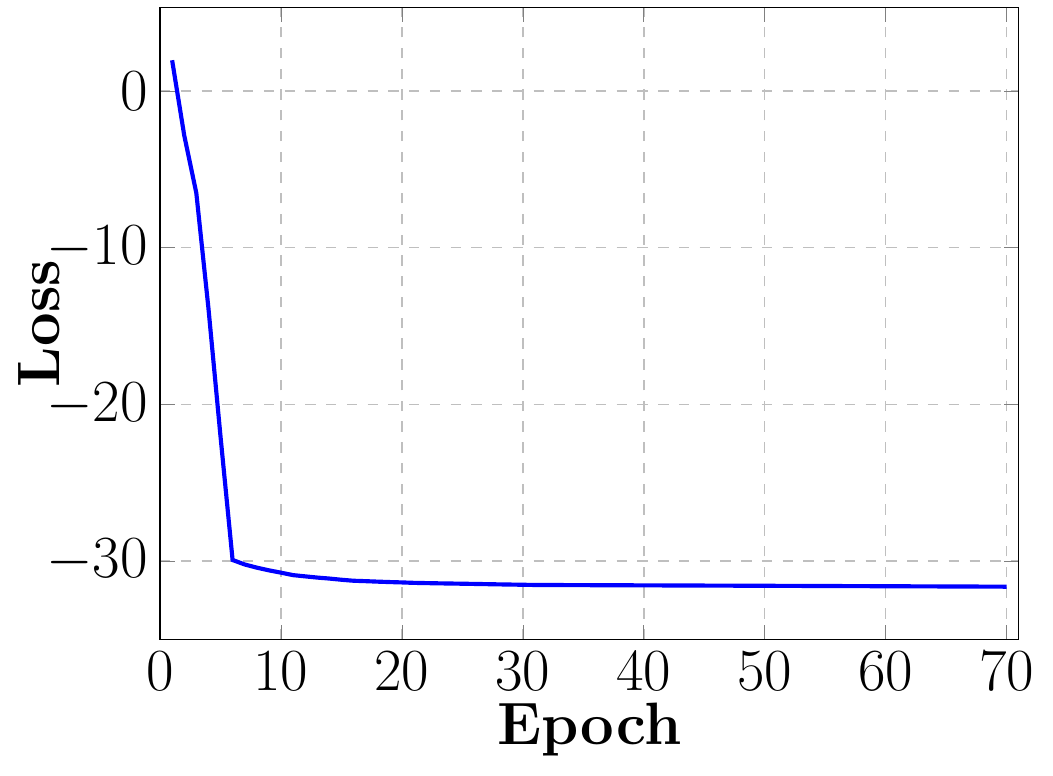}%
\label{fig:loss_vs_epoch_iptv}}
\vspace{0.01in}\\
\subfloat[]{\includegraphics[height=1.35in,width=1.75in]{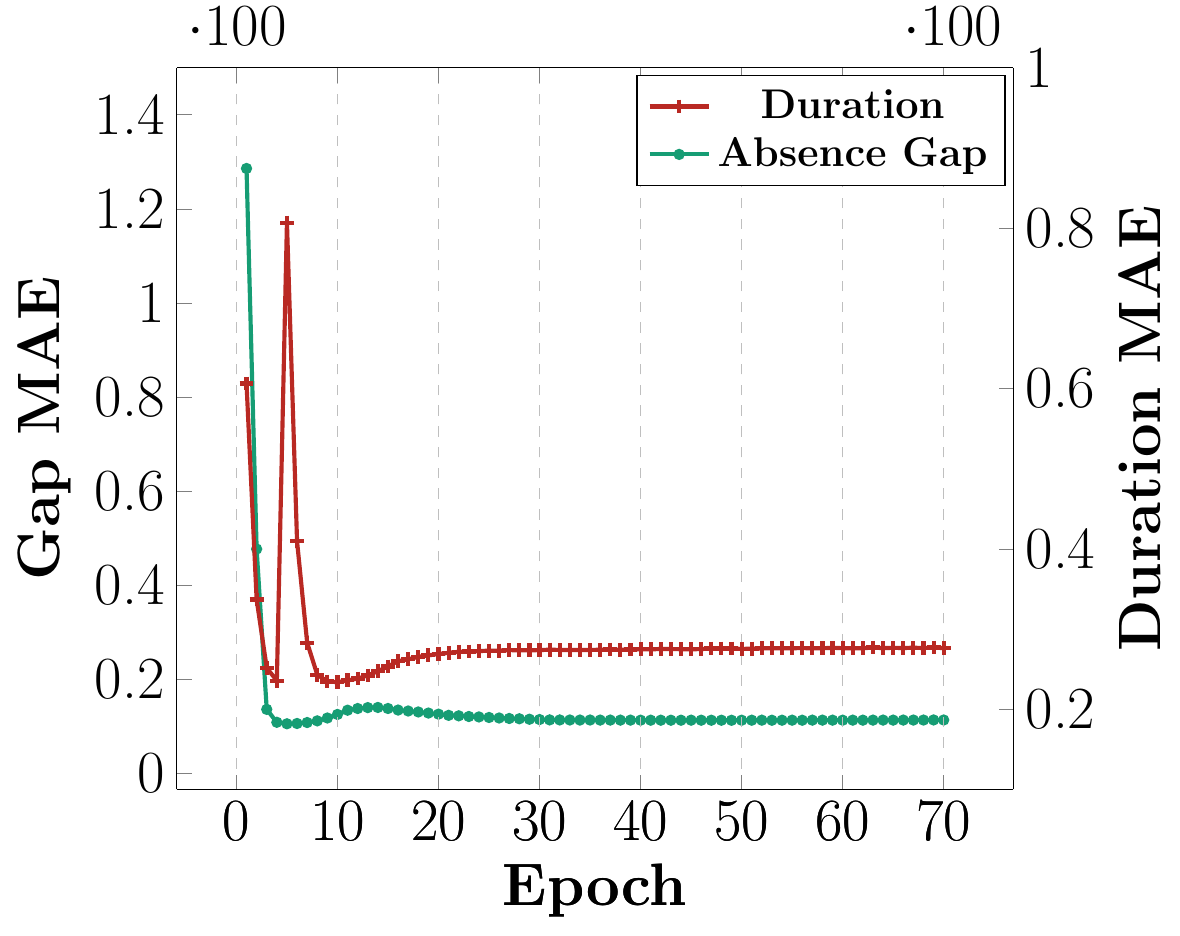}%
\label{fig:time_mse_vs_epoch_tianchi}}
\hspace{0.01in}%
\subfloat[]{\includegraphics[height=1.35in,width=1.75in]{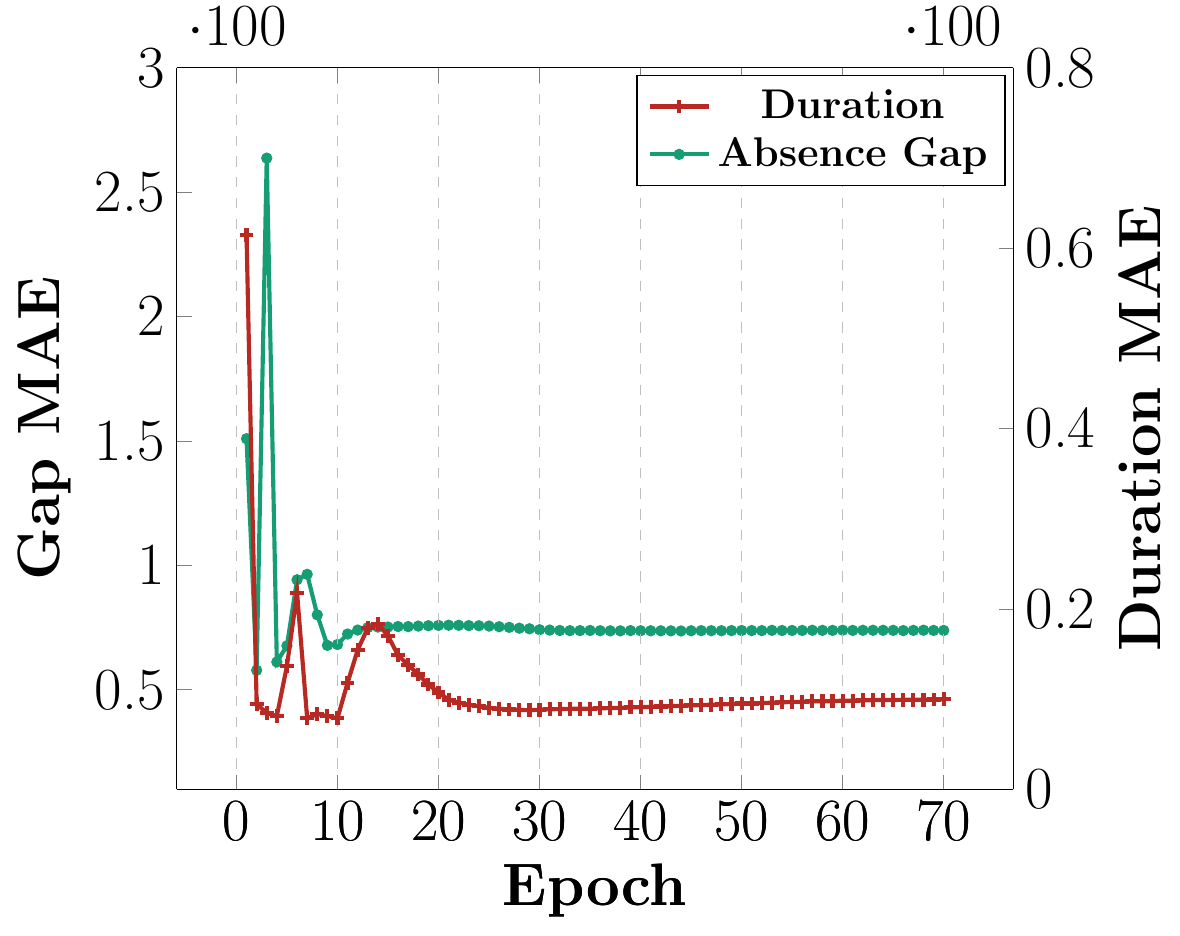}%
\label{fig:time_mse_vs_epoch_lastfm}}
\hspace{0.01in}%
\subfloat[]{\includegraphics[height=1.35in,width=1.75in]{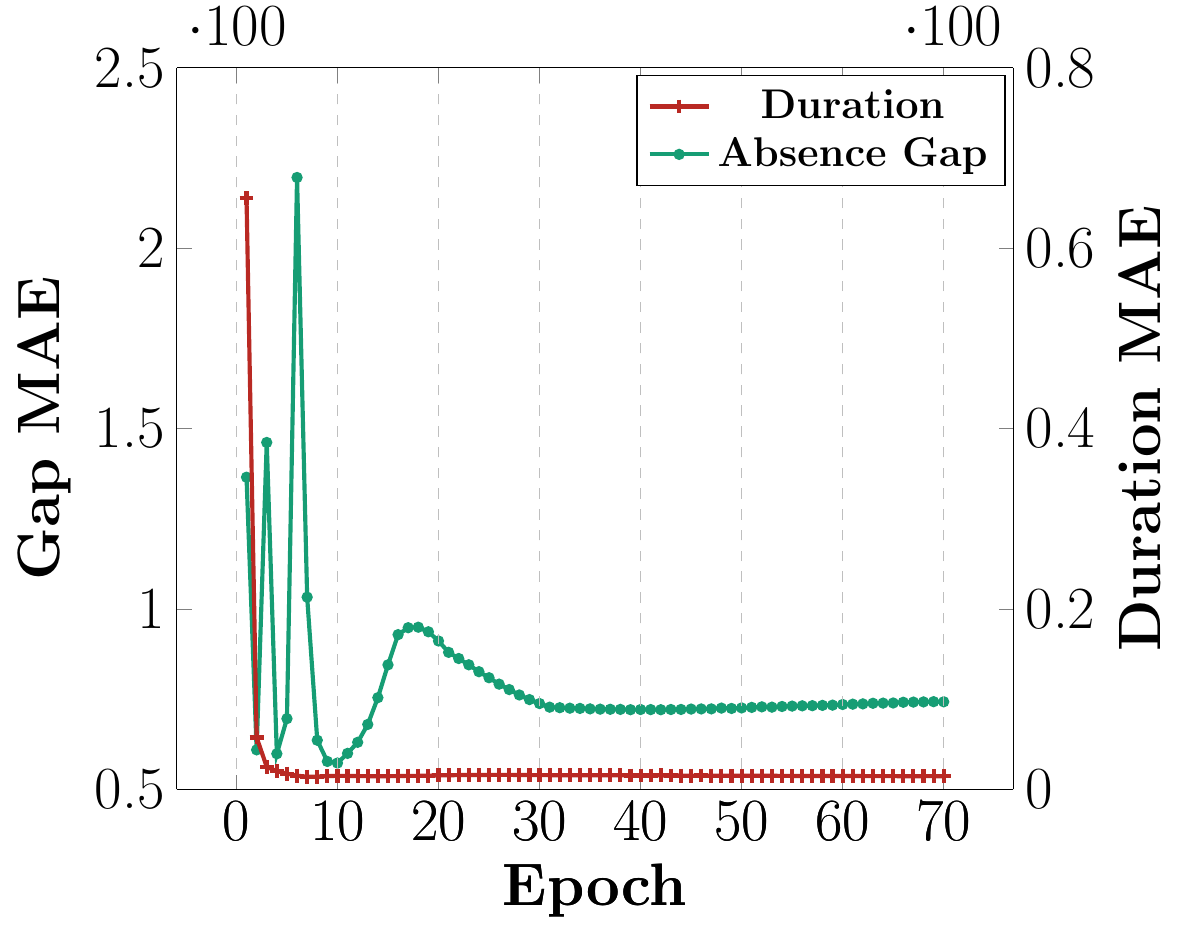}%
\label{fig:time_mse_vs_epoch_foursquare}}
\hspace{0.01in}%
\subfloat[]{\includegraphics[height=1.35in,width=1.75 in]{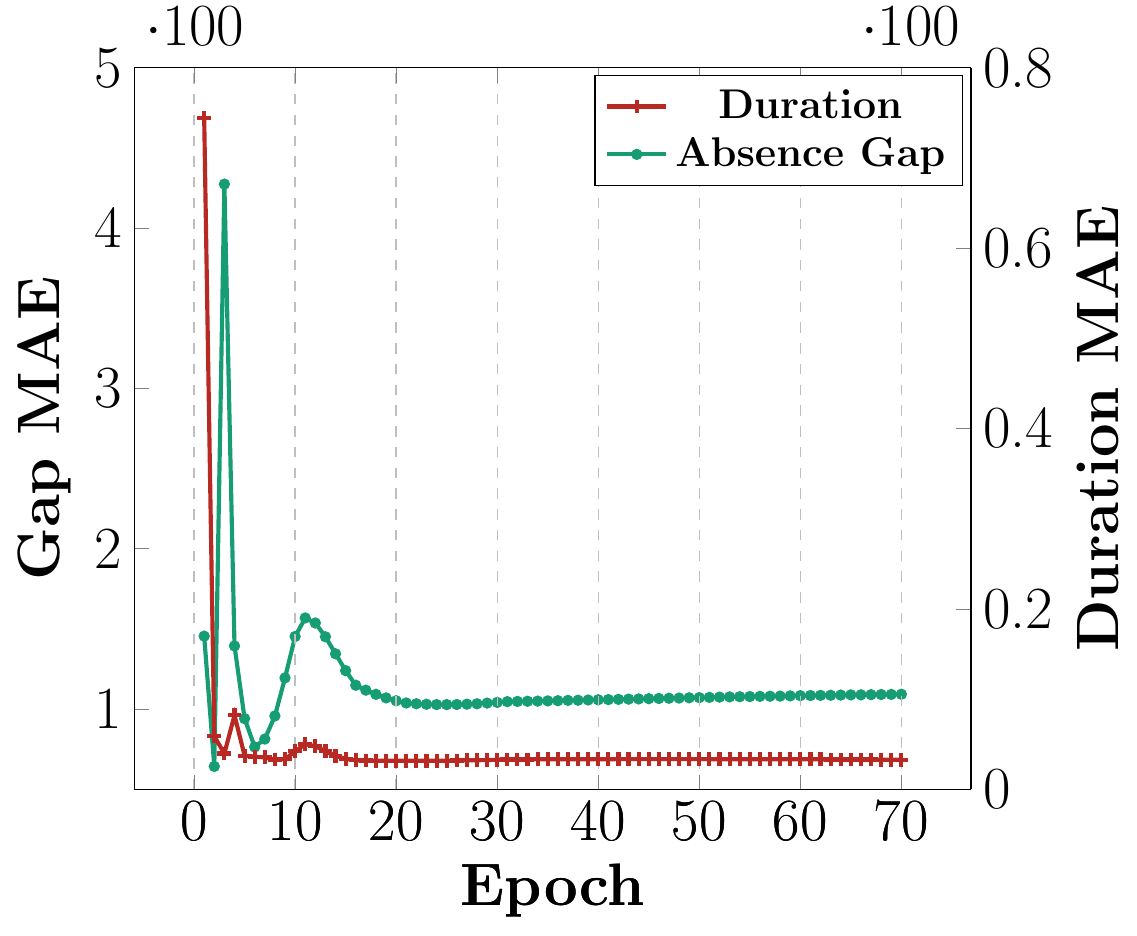}%
\label{fig:time_mse_vs_epoch_iptv}}
\caption{The convergence of proposed method. First row presents the \textbf{Loss} on train data against different epochs for (a) Tinachi, (b) Last.fm, (c) Foursquare, and (d) IPTV datasets. Second row, presents the \textbf{MAE of absence prediction} and \textbf{MAE of session duration prediction} on train data against different epochs. The threshold for session extraction is set to $1$ hour.
}
\label{fig:convergence_study} 
\end{figure*}

\section{Experiments}
We evaluate the performance of ChOracle on large-scale real world datasets. We first describe the competitor baselines and the datasets. Then, we study the convergence of proposed learning algorithm. Next, we evaluate the performance of the proposed method in predicting the absence gap of users and their future session durations. Finally, we study the impact of session based approach on the performance of proposed method.
\subsection{Competitor baselines}
We cast the churn prediction problem into a return time prediction problem. Therefore, we evaluate performance of the proposed method in both predicting the next absence gap and next session duration. However, none of the previous works have tried to do this type of evaluation. Only some works have tried to predict the return time of users to the items. However, the main goal of many of these works are neither churn prediction nor return time prediction to the system, but some of them can be modified to predict user return times to the system for comparisons. Therefore, to evaluate the performance of proposed method in predicting the next return time to the system, we compared ChOracle with the following models.
\begin{itemize}
	\item \textbf{RMTPP}\cite{RMTPP}. This method tries to model the intensity function of a temporal point process using recurrent neural networks. It learns a general intensity function that can predict next event time and marker. Although its primary intention is not for churn prediction and its markers have discrete finite values, but we used its temporal part for return time prediction to the system. 
	\item \textbf{NSR}\cite{NeuralSurvival}. The Neural Survival Recommender is designed for recommendation purposes and tries to model the return times of users to the items. But, since it also tries to model the temporal point process using RNNs based on a different approach, we also compared its performance in return time prediction with the proposed method.
\end{itemize}

As discussed in the related works section, the authors in \cite{kapoor2014hazard} have also tried to model the user return times to service, however their method is application dependent and uses detailed specific features that makes it only applicable to music streaming services. Since in our experiments we used datasets from different general domains and only use high level user interaction data such as session duration and absence gap, we did not include this method in the competitor baselines.

\textbf{Experimental Setup.}\\
For implementation of deep networks, we used TensorFlow \cite{TensorFlow}, a scalable deep learning library in python. To be fair, we trained all three methods that are based on neural networks (ChOracle, RMTPP, NSR) with 70 iterations on the train data with the learning rate of 0.001. We used a single LSTM layer with 300 hidden units. We did not use any embedding for the proposed method, RMTPP, and NSR. This will impact the results but will help us to evaluate the performance of methods without any help of embeddings. We selected 80\% of users as the training set and the remaining 20\% as the test set. We set the session length threshold to 1 hour for all datasets and experiments, except when explicitly stated. More information about the session length is provided in related subsections of the results. 

In addition, we set $w^t=0$ in Eq. \ref{eq:lambda}, in implementation of ChOracle. For simplicity in calculations of the KL divergence, instead of using $z$ we used $logit(z)$ which has a Gaussian distribution.


\subsection{Dataset Description}

We evaluated the performance of ChOracle on four real datasets from different domains; \emph{Last.fm}, \emph{Tianchi Alibaba Mobile Commerce}, \emph{Foursquare}, and \emph{IPTV}. All datasets contain timestamped actions which make them appropriate benchmarks for comparing the proposed method to the other state-of-the-art methods.

\textbf{Last.fm.}
This dataset contains the music listening logs of 1200 users and 1000 artists. There are around 418K events in total which spans a period of 6 months. 

\textbf{Tianchi Mobile Data.} It contains the user interactions with items in Alibaba's mobile M-Commerce platform~\cite{alibabaDataSet}. The dataset includes four behavior types: click, collect, add-to-cart, and payment. We only considered  the click events. Our data contains roughly 1000 users, 2100 items, and a total of 1.2M events.

\textbf{Foursquare Data.} This dataset contains Foursquare users' check-ins in London~\cite{LondonDataset}. We selected the active users with more than 30 check-ins and the venues with more than 50 check-ins which resulted in 67K check-ins of 890 users in London, which spans from Mar. 2011 to Sep. 2011.

\textbf{IPTV Data.} This dataset contains the users' history of watching TV programs on online TV streaming services~\cite{CoevolutionaryRecommenderSystem}. The dataset contains 7100 users and 436 TV programs. The dataset contains 2.4M events and spans a period of 11 months.


\begin{figure*}[!t]
\centering
\subfloat[Tianchi]{\includegraphics[width=1.75in]{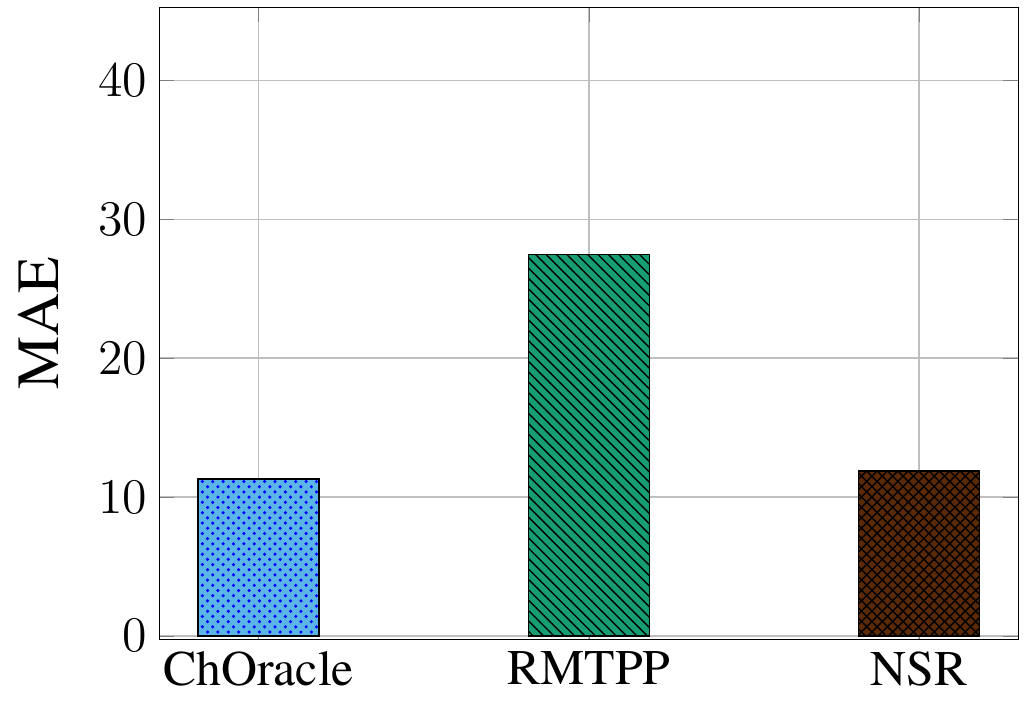}%
\label{fig:gap_mae_tianchi}}
\hspace{0.01in}%
\subfloat[Last.fm]{\includegraphics[width=1.75in]{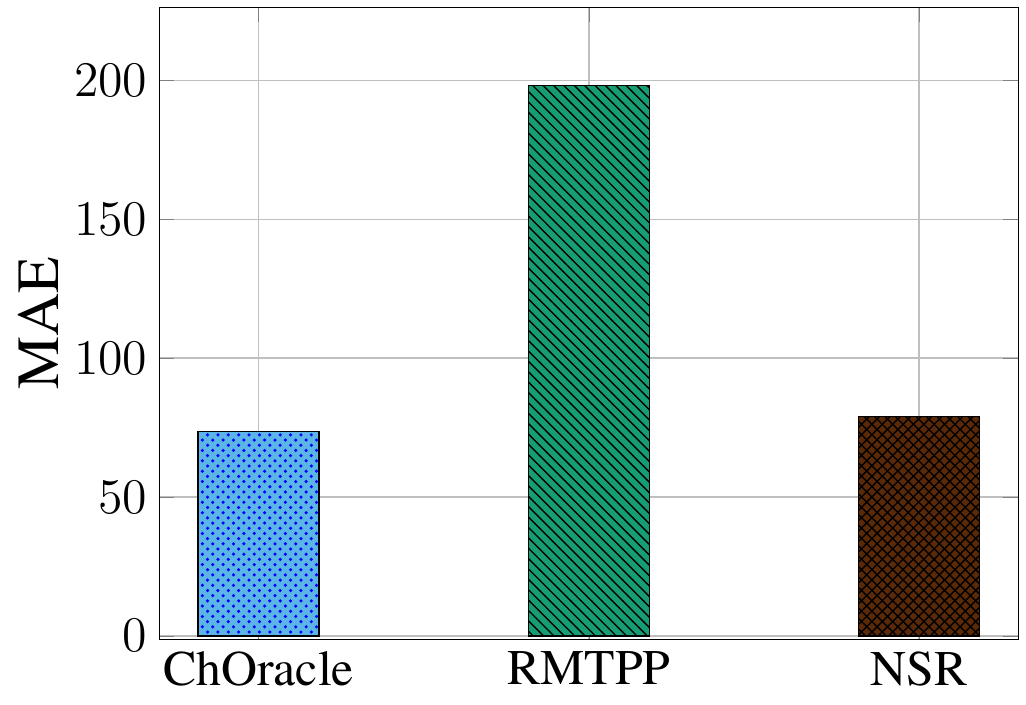}%
\label{fig:gap_mae_lastfm}}
\hspace{0.01in}%
\subfloat[Foursquare]{\includegraphics[width=1.75in]{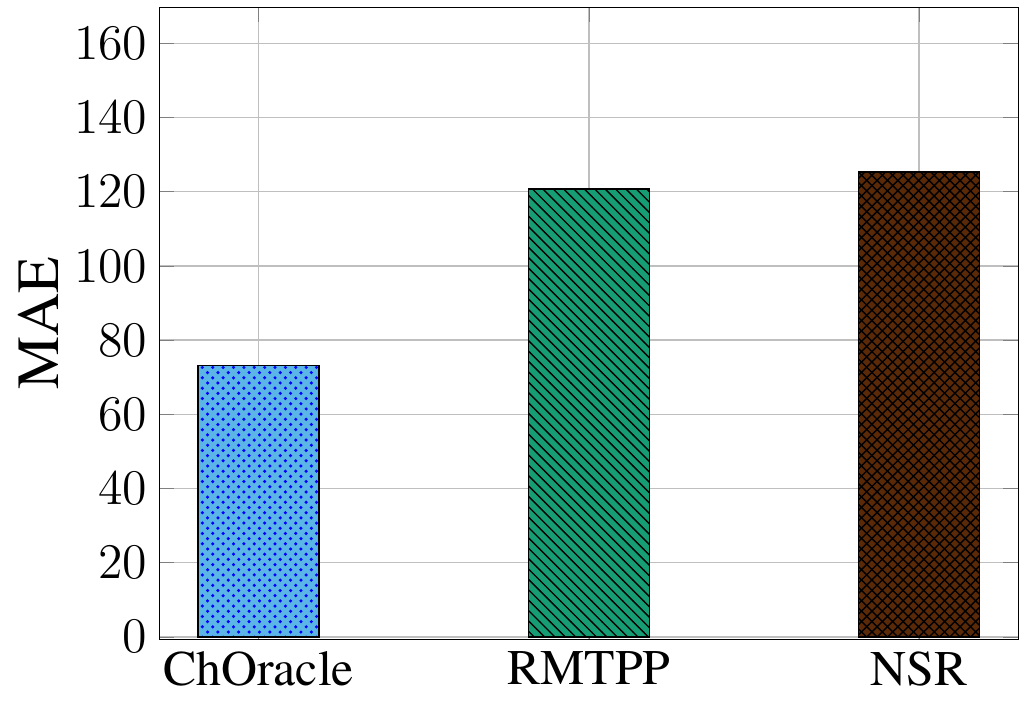}%
\label{fig:gap_mae_foursquare}}
\hspace{0.01in}%
\subfloat[IPTV]{\includegraphics[width=1.75in]{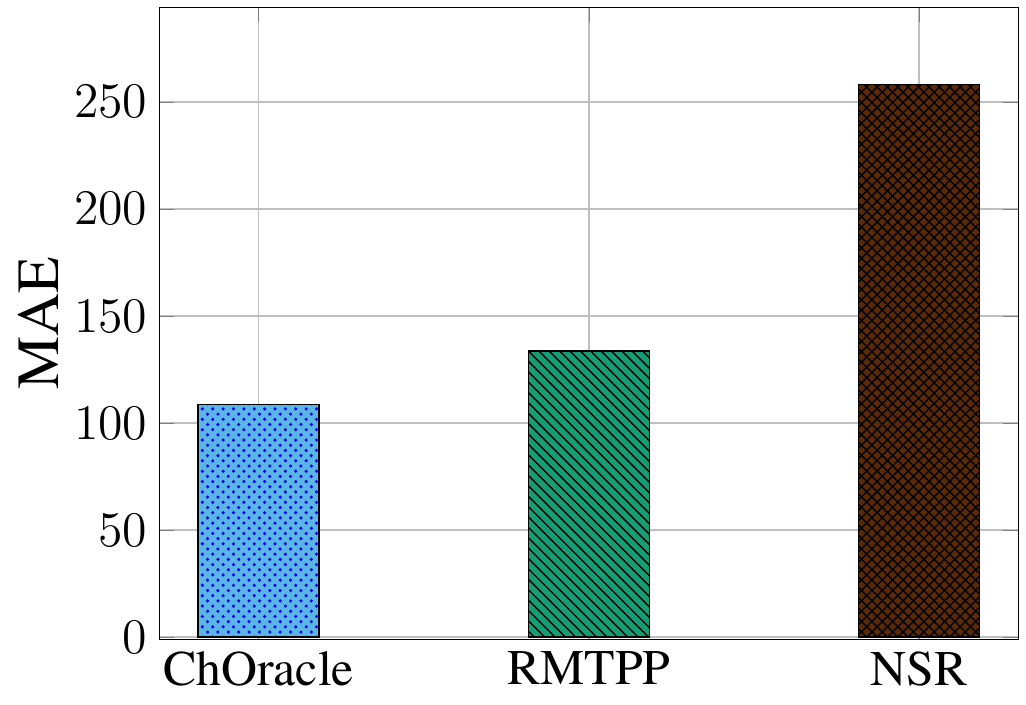}%
\label{fig:gap_mae_iptv}}
\hspace{0.01in}%
\caption{ Performance of different methods in predicting the next absence gap of users on test data for (a) Tianchi, (b) Last.fm, (c) Foursquare, and (d) IPTV datasets.
}
\label{fig:absence_gap_results}
\end{figure*}
\begin{figure*}[t]
\centering
\subfloat[Tianchi]{\includegraphics[width=1.85in]{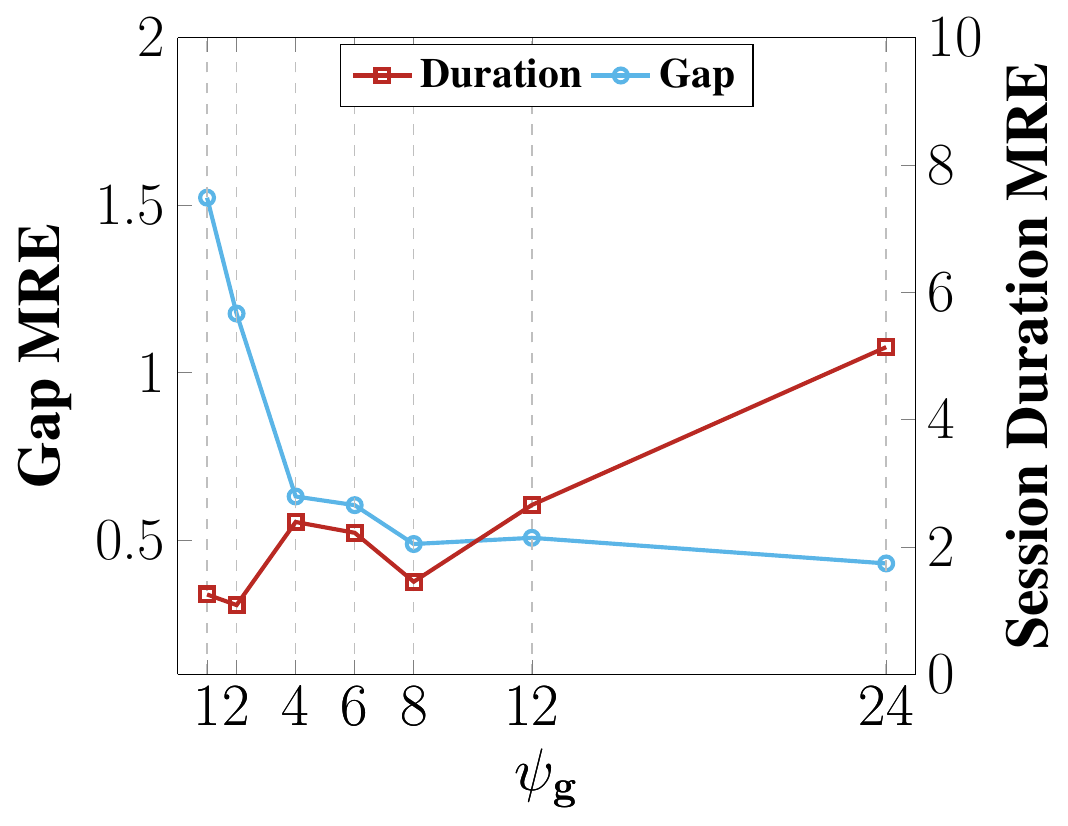}%
\label{fig:MAEOverIterations_Synth}}
\hspace{0.01in}%
\subfloat[Last.fm]{\includegraphics[width=1.75in]{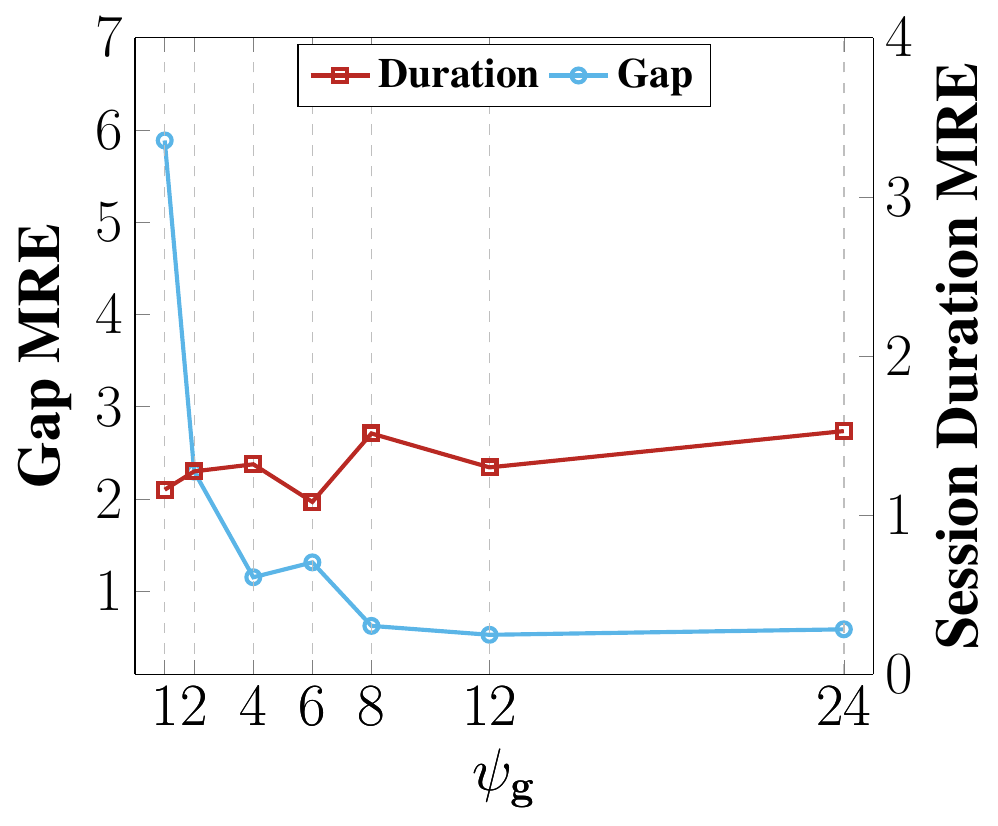}%
\label{fig:MAEOverEvents_Synth}}
\hspace{0.01in}%
\subfloat[Foursquare]{\includegraphics[width=1.75in]{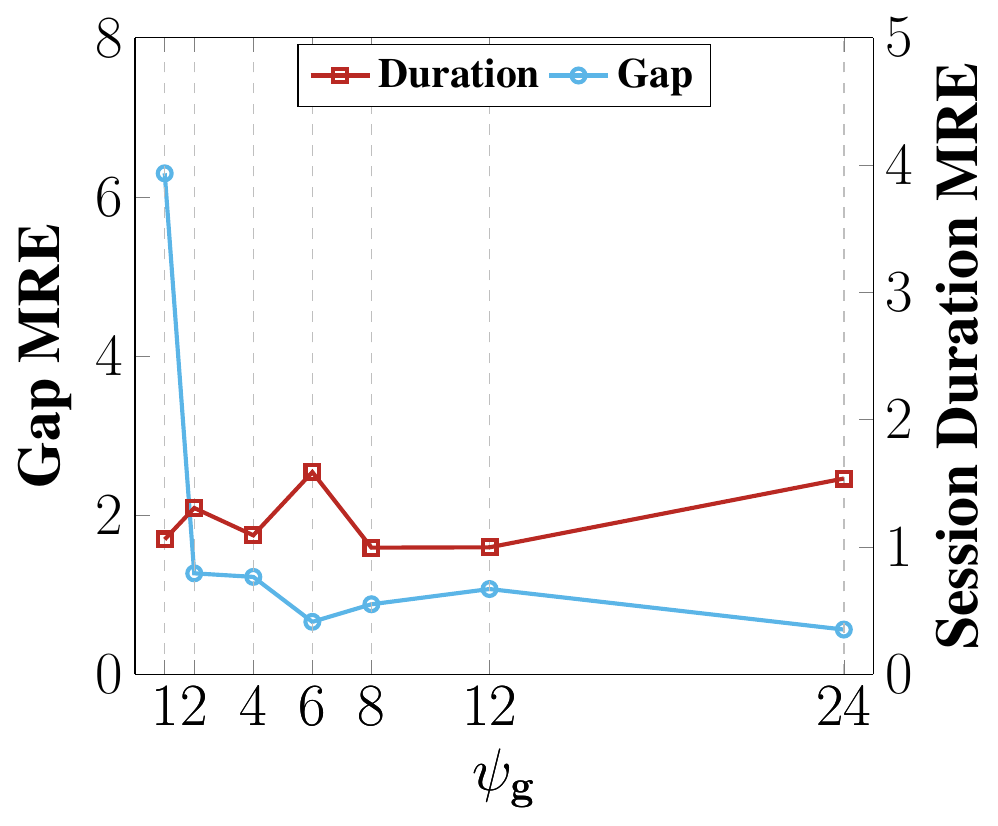}%
\label{fig:QQPlot_Synth}}
\hspace{0.01in}%
\subfloat[IPTV]{\includegraphics[width=1.75in]{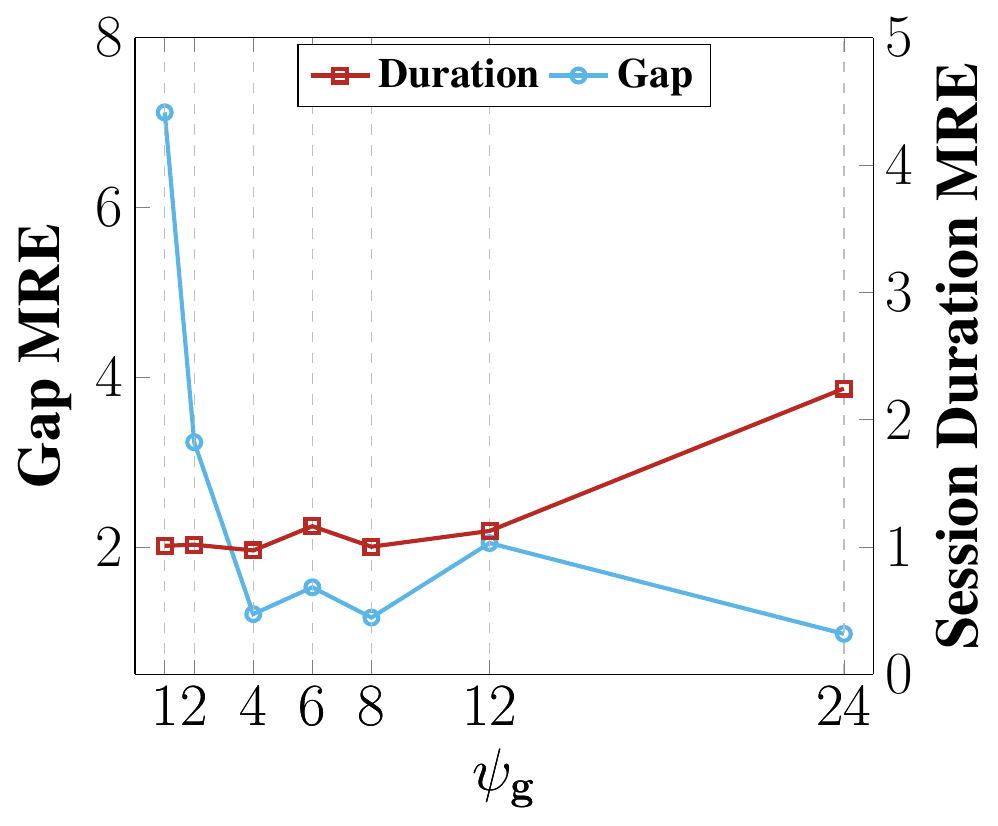}%
\label{fig:Scalability_Synth}}
\caption{Impact of session threshold $\psi_g$ on the results for a) Tianchi, b) Last.fm, c) Foursquare, and d) IPTV datasets. \textbf{blue} plot shows the MRE of absence gap prediction, \textbf{red} plot shows the MRE of session length prediction.}
\label{fig:session_length_study}
\end{figure*}

\subsection{Results}

\textbf{Convergence Analysis} \\
In fig. \ref{fig:convergence_study} we illustrate the convergence behavior of the proposed method. We investigated how the \textbf{Loss}, and \textbf{MAE} changes through epochs on the train data.
First row presents the Loss against number of epochs for different datasets. The presented Loss is the negative of ELBO introduced in Eq. \ref{eq:ELBO} divided by the number of events; i.e. negative of average ELBO per event. Since we try to maximize the ELBO as the objective function, we expect that the its negative decreases with an increase in the number of epochs. As it can be clearly seen from the first row, with an increase in the number of epochs the loss decreases in all data sets. It cab be also noticed that after about $20$ epochs, the loss converges to a fixed value which is a sign of fast convergence in all datasets. 
As the main goal of ChOracle is to predict when the user will comeback to the system, in the second row, we also presented the MAE of absence gap, and session duration prediction on the train data for different datasets. Minimizing the absence gap, and session duration prediction error is not the main objective of the optimization, but as the second row shows, minimizing the loss will result in minimizing the absence gap, and session duration prediction error in all datasets. 
As it can be seen, the absence gap prediction error converges rapidly for Tianchi and Last.fm datasets while it takes more epochs for Foursquare and IPTV datasets to converge. In the contrary, the session duration prediction error converges rapidly for Foursquare and IPTV datasets while it took more epochs to converge for Tianchi, and Last.fm datasets.
\\\\\textbf{Absence Gap Prediction}\\
In fig. \ref{fig:absence_gap_results} we illustrate the performance of different methods in predicting the next absence gap of users on the test data. As depicted, the proposed method outperforms the competitors in all datasets. 
The NSR performance is close to the proposed method in Tianchi and Last.fm datasets (Figs. \ref{fig:gap_mae_tianchi}, \ref{fig:gap_mae_lastfm}). As we mentioned previously, the NSR method suffers from high complexity and requires more computational power to reach the desirable results. It learns a vector of length $4320$ for each event that is the intensity function of a temporal point process for about $180$ days. The Foursquare dataset has only $800$ users and many of its sessions contain just 1 event. Hence, NSR do not perform well for the Foursquare dataset. IPTV dataset contains about $5000$ users and about 1M events. When we tried to fit NSR on the IPTV dataset by using a simulation server with 12GB of GPU, we faced out of memory (OOM) errors, since the data did not fit in the memory. To resolve this issue, we reduced the length of prediction to only $500$ future hours and hence it could not well predict the future events. This is depicted in the Fig. \ref{fig:gap_mae_iptv}. 
The RMTPP method, which only uses RNNs to model the timing of events, can not well describe the latent patterns that govern the temporal dynamics of the events. It only performs close to the proposed method for the IPTV dataset, which has a lot of training data. However, for datasets with fewer data, it cannot well describe the patterns governing the temporal dynamics of data. Instead, the proposed method which uses RNNs to define the intensity function of temporal point process and incorporates the latent variables into the RNNs, can well describe the latent patterns that govern the temporal dynamics of events. Moreover, it does not suffer from high computational complexity.
\\\\\textbf{Session Duration Prediction}\\
 We also studied performance of the proposed method in predicting the next session duration. We chose the number of events in the session as its duration. For example, the session duration for Last.fm dataset is the number of songs the user listened in a session, and the session duration for Tianchi, Foursquare, and IPTV is the number of products the user clicked, the number of checkins the user did, and the number of tv programs the user watched in the session, respectively. The results is presented in Table. \ref{table:session_duration_results}. It should be noted that our primary goal is not to exactly predict the session duration and we only use it for better prediction of the future return times. As the results demonstrate, the proposed method performs better on the Foursquare and IPTV datasets compared to Tianchi and Last.fm. Since the session duration in Foursquare and IPTV datasets is less than the Tianchi and Last.fm datasets, the average error is also expected to be less for those datasets. The Mean Relative Error (MRE) is the same over all datasets which shows that relative performance do not change over different datasets, and the differences in MAE is because of different scales in session duration of different datasets.
\begin{table}[!t]
\renewcommand{\arraystretch}{1.45}
\caption{Session Duration Prediction Results}
\label{table:session_duration_results}
\centering
\begin{tabular}{|c||c|c|c|c|}
\hline
\textbf{Dataset} & Tianchi & Last.fm & Foursquare & IPTV\\
\hline
\textbf{MAE}& 23.48 & 7.57 & 1.43 & 3.13\\
\hline
\textbf{MRE}& 1.25 & 1.16 & 1.06 & 1.00\\
\hline
\end{tabular}
\end{table}
%
\\\\\textbf{Impact of Session Threshold on the performance}\\
There is no explicit notion of session in the datasets we used for our experiments. Therefore, we manually segmented the events into sessions. To this end, we selected a minimum gap ($\psi_g$) and if the gap between two consecutive events ($i$, $i+1$) is less than the threshold ($t_{i+1}-t_i < \psi_g$) we assume that they belong to the same session. We used this method to create session based events. The choice of threshold $\psi_g$ is application dependent and may affect the results.
 
Fig. \ref{fig:session_length_study} shows the impact of session length threshold $\psi_g$ on performance of the proposed method both in predicting the next absence gap and next session duration. To this end, we plotted the \textbf{MRE} of these two metrics against the session threshold. As illustrated in this figure, with increasing the threshold $\psi_g$, the MRE of absence gap prediction decreases for all datasets. Because with increasing the threshold the resulting sessions will be longer and the gaps between sessions increases, and as a result the relative error decreases. In the contrary, with an increase in the threshold there is no significant change in the MRE of next session predictions, and the results do not change dramatically. 

\section{Conclusions}
In this work, we presented a novel framework, \emph{ChOracle}, for churn prediction in online services. 
ChOracle models the user absence gaps and session durations, by extending temporal point processes. 
In order to model general temporal intensities, ChOracle uses recurrent neural networks to define the intensity function of temporal point processes. Therefore, it can adaptively model different intensity functions.
We also incorporated latent random variables into the hidden states of RNN, which adds more expressive power to the model and enables ChOracle to deal with highly structured data. 
The last but not the least, we derived a variational lower bound as the objective function. By maximizing this objective function we can learn all the parameters by using back propagation through time (BPTT). 
Experiments on real world datasets demonstrate the superiority of the proposed framework over state-of-the-art methods.

For future work, one may use more specific data about user sessions to improve the predictive performance of the proposed method. We also would like to investigate defining churners based on predicted absence gaps in a real world scenario. Another interesting venue for future work is utilizing generative adversarial neural networks (GAN) \cite{xiao2017wasserstein} to model the intensity functions of temporal point processes. 
\newpage
\appendices
\section{Derivation of ELBO}

For simplicity, we derive the ELBO for event sequence of a single user $u$, since it is straightforward to generalize the results to many users. 
\begin{align} \label{eq:elbo_app}
&\mathcal{L}\Big(\mathcal{S}^u(T)\Big) =
\log \int P \Big(\mathcal{S}^u(T),\mathcal{Z}^u(T) \Big)d\mathcal{Z} \nonumber\\
&=\log \int p(g_{1:T}, d_{1:T},z_{1:T})dz_{1:T} \nonumber\\
&= \log \int q(z_{1:T}|g_{1:T},d_{1:T}) \frac{p(g_{1:T}, d_{1:T},z_{1:T})}{q(z_{1:T}|g_{1:T},d_{1:T})} dz_{1:T} \nonumber\\
&\geq \int q(z_{1:T}|g_{1:T},d_{1:T}) \log \left( \frac{p(g_{1:T}, d_{1:T},z_{1:T})}{q(z_{1:T}|g_{1:T},d_{1:T})} \right) dz_{1:T} \nonumber\\
&= \int  q(z_{1:T}|g_{1:T},d_{1:T}) \times \nonumber\\
&\log\left(\frac{\prod_{i=1}^T p(z_i| z_{<i},g_{<i}, d_{<i})p(g_i|z_{<i},g_{<i}, d_{<i})p(d_i|z_{<i},g_{<i}, d_{<i})}{\prod_i q(z_i|z_{<i},g_{\leq i},d_{\leq i})} \right) dz
\end{align}
We can decompose the above equation into two parts:
\begin{align}
	A &= \int  q(z_{1:T}|g_{1:T},d_{1:T}) \times \nonumber\\
&\sum_i \Big( \log p(g_i|z_{<i},g_{<i}, d_{<i})+ \log p(d_i|z_{<i},g_{<i}, d_{<i}) \Big) dz \nonumber\\ 
&= \mathbf{E}_{q(z_{1:T}|g_{1:T},d_{1:T})} \Bigg\lbrack 
\sum_i
\log p(g_i|z_{<i},g_{<i}, d_{<i})+ \log p(d_i|z_{<i},g_{<i}, d_{<i})
\Bigg\rbrack 
\end{align}

\begin{align}
	B &= \int  q(z_{1:T}|g_{1:T},d_{1:T}) \times \nonumber\\
&\sum_i \Big( \log p(z_i| z_{<i},g_{<i}, d_{<i})-\log q(z_i|z_{<i},g_{\leq i},d_{\leq i}) \Big) dz \nonumber\\
&=\sum_i \int q(z_{\leq i}|g_{\leq i},d_{\leq i}) \times \nonumber\\
& \Big( \log p(z_i| z_{<i},g_{<i}, d_{<i})-\log q(z_i|z_{<i},g_{\leq i},d_{\leq i}) \Big) dz \nonumber\\
&=-\sum_i \int q(z_{< i}|g_{< i},d_{< i}) \times \nonumber\\
 &\mathbf{KL}\Big[q(z_i|z_{<i},g_{\leq i},d_{\leq i})\parallel p(z_i| z_{<i},g_{<i}, d_{<i})\Big] dz
 \nonumber\\
&=-\mathbf{E}_{q(z_{1:T}|g_{1:T},d_{1:T})} \Bigg\lbrack
\nonumber\\
&\sum_i
\mathbf{KL}\bigg[q(z_i|z_{<i},g_{\leq i},d_{\leq i})\parallel p(z_i| z_{<i},g_{<i}, d_{<i})\bigg]
\Bigg\rbrack 
\end{align}

and hence, ELBO will be as follows:
\begin{align}
	&\mathcal{L}\Big(\mathcal{S}^u(T)\Big) = \nonumber\\
	& \mathbf{E}_{q(z_{1:T}|g_{1:T},d_{1:T})} \Bigg\lbrack 
	\sum_i \bigg[
	\log p(g_i|z_{<i},g_{<i}, d_{<i}) \nonumber\\
	&+\log p(d_i|z_{<i},g_{<i}, d_{<i}) \nonumber\\
	&+\mathbf{KL}\Big[q(z_i|z_{<i},g_{\leq i},d_{\leq i})\parallel p(z_i| z_{<i},g_{<i}, d_{<i})\Big] \bigg]
	\Bigg\rbrack \nonumber\\
\end{align}
In our experiments, to calculate the expectation with respect to the $q(.)$ distribution, we draw $L$ samples from it and then compute the ELBO as follows:
\begin{align}
	&\mathcal{L}\Big(\mathcal{S}^u(T)\Big) = \nonumber\\
	& \frac{1}{L}\sum_{l=1}^L\Bigg\lbrack 
	\sum_i \bigg[
	\log p(g_i|z^l_{<i},g_{<i}, d_{<i}) \nonumber\\
	&+\log p(d_i|z^l_{<i},g_{<i}, d_{<i}) \nonumber\\
	&+\mathbf{KL}\Big[q(z^l_i|z^l_{<i},g_{\leq i},d_{\leq i})\parallel p(z^l_i| z^l_{<i},g_{<i}, d_{<i})\Big] \bigg]
	\Bigg\rbrack \nonumber\\
	&\qquad \mathbf{s.t.} \quad
	z^l_{1:T} \sim q(z_{1:T}|g_{1:T},d_{1:T})
\end{align}

%
%


{	
\bibliographystyle{IEEEtran}
\bibliography{Choracle}
}
\vspace*{-2\baselineskip}
\begin{IEEEbiography}[{\includegraphics[width=1.1in,height=1.3in,clip,keepaspectratio]{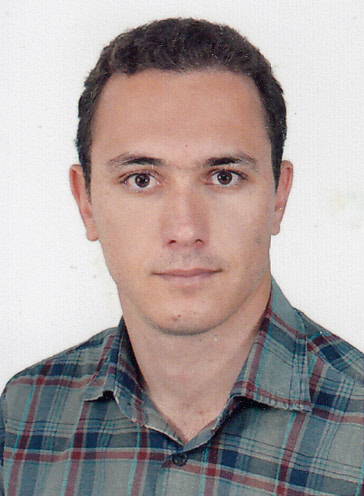}}]{Ali Khodadadi}
received his B.Sc. and the M.Sc. degrees in information technology engineering from Sharif University of Technology, Tehran, Iran, in 2010 and 2012, respectively. He is currently a Ph.D. candidate in the Department of Computer Engineering at Sharif University of Technology.
His current research interests include machine learning with application in social and complex networks including inferring networks of diffusion, Bayesian community detection, multilayer network analysis, and user activity modeling.
\end{IEEEbiography}
\vspace*{-2\baselineskip}
\begin{IEEEbiography}[{\includegraphics[width=1.1in,height=1.3in,clip,keepaspectratio]{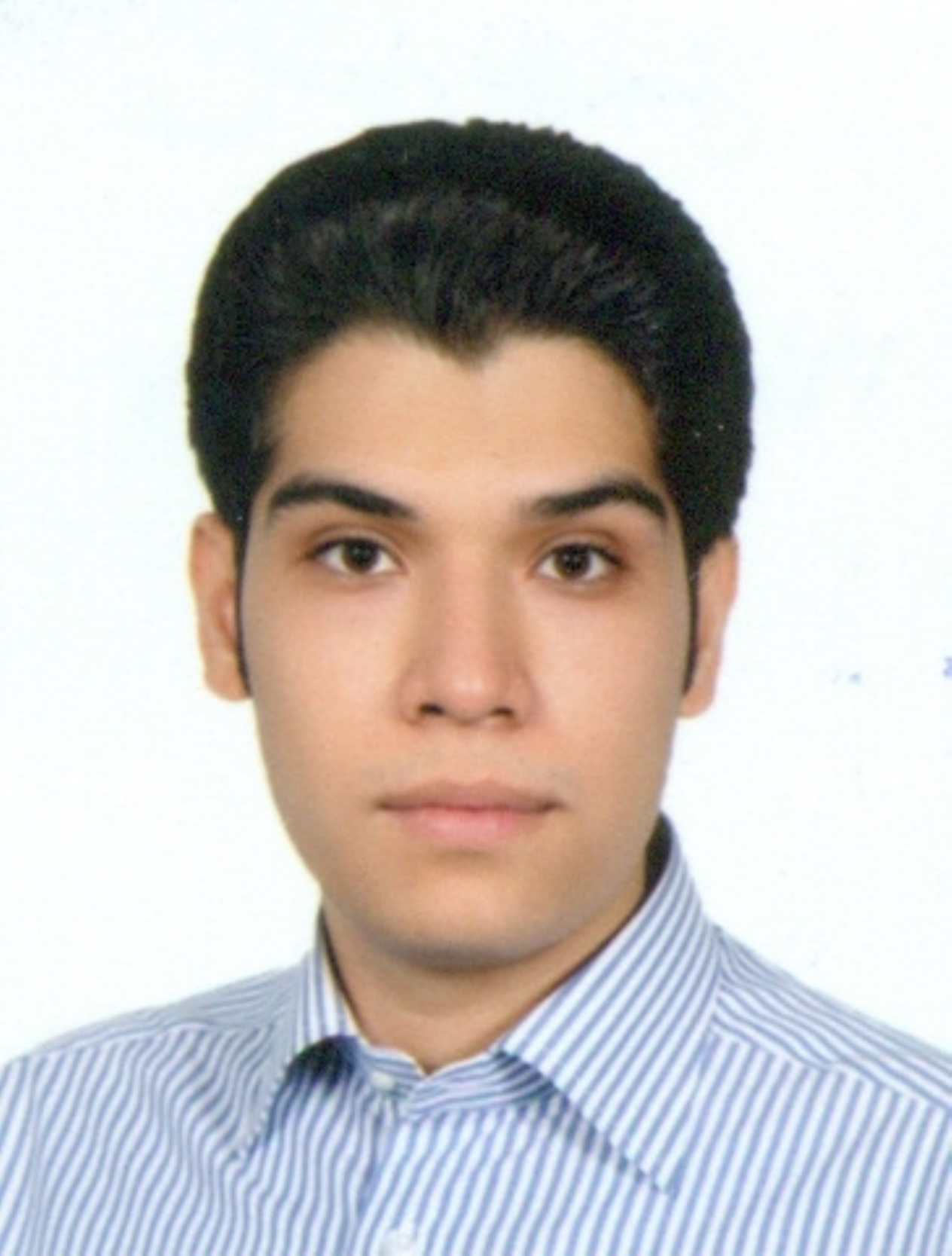}}]{Seyed Abbas Hosseini}
received his B.Sc. in Software Engineering and M.Sc. in Artificial Intelligence from Sharif University of Technology, Tehran, Iran, in 2012 and 2014, respectively. He is currently a Ph.D. candidate in the Department of Computer Engineering at Sharif University of Technology.
Since 2014 he works as a technical consultant at Intelligent Data Processing Group (PegahTech) and is an adjunct lecturer at Sharif University of Technology. His current research interests include modeling marked events, user profiling and their applications in recommendation and online advertising.
\end{IEEEbiography}

\vspace*{-2\baselineskip}
\begin{IEEEbiography}[{\includegraphics[width=1.1in,height=1.3in,clip,keepaspectratio]{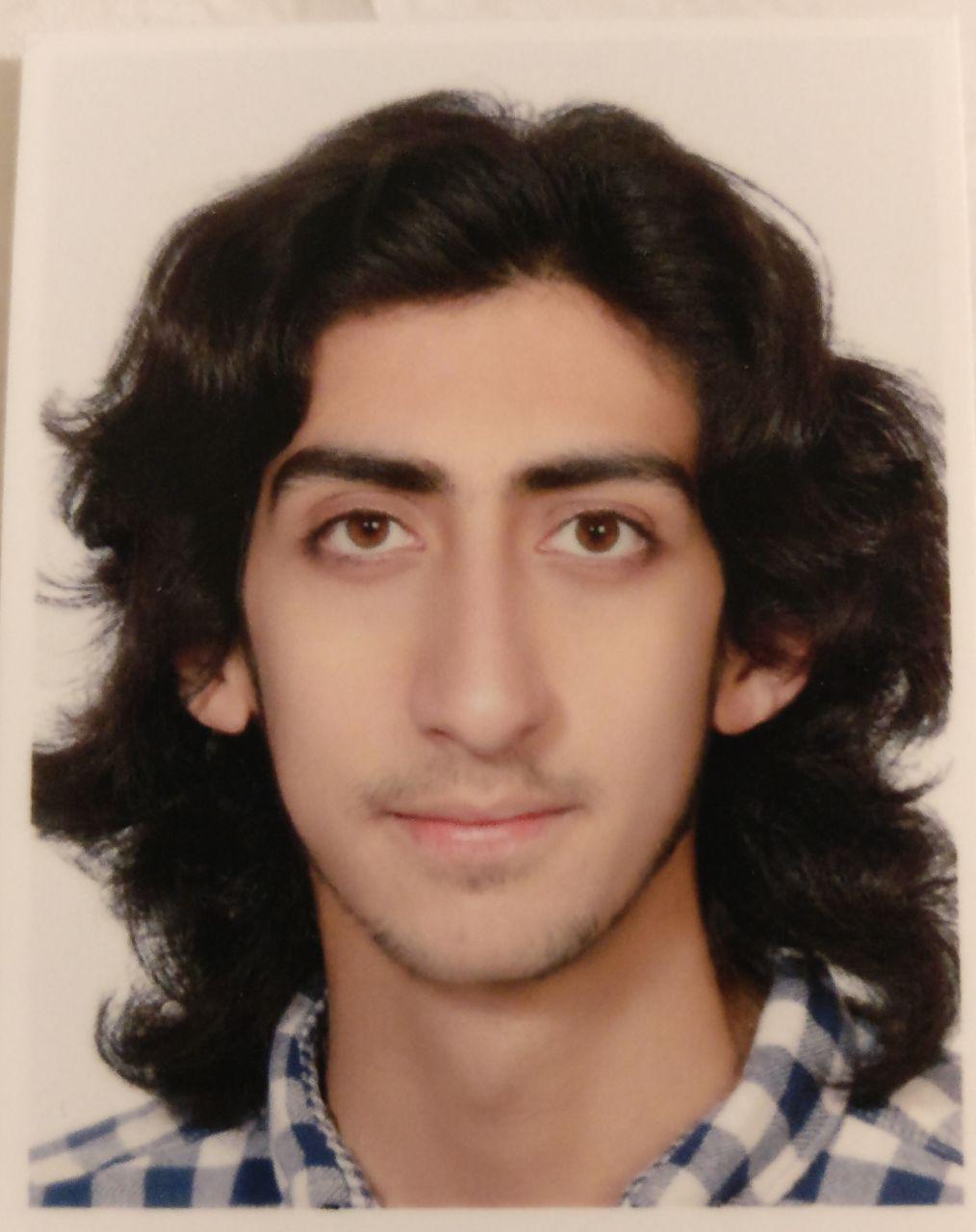}}]{Ehsan Pajouheshgar}
is currently a B.Sc. student in the department of Computer Engineering at Sharif University of Technology, Tehran, Iran. In summer 2018 he was an intern at IST Austria under the supervision of Christoph H. Lampert. He also received a national gold medal in physics Olympiad in 2014. His current research interests include Deep Learning, specially sequence modeling and sequence generation, and their applications in Computer Vision and Natural Language Processing.
\end{IEEEbiography}

\vspace*{-2\baselineskip}
\begin{IEEEbiography}[{\includegraphics[width=1.1in,height=1.3in,clip,keepaspectratio]{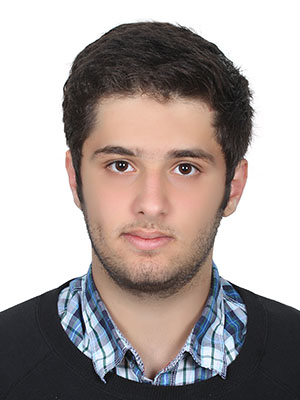}}]{Farnam Mansouri}
is currently a B.Sc. Student in software engineering at the department of Computer Engineering at Sharif University of Technology, Tehran, Iran. He had a fellowship in machine learning at Max Planck Institute of Software Systems. His current research interests include using deep learning in medical imaging, information theory, structure learning, and learning complexity.
\end{IEEEbiography}

\vspace*{-4\baselineskip}
\begin{IEEEbiography}[{\includegraphics[width=1.1in,height=1.3in,clip,keepaspectratio]{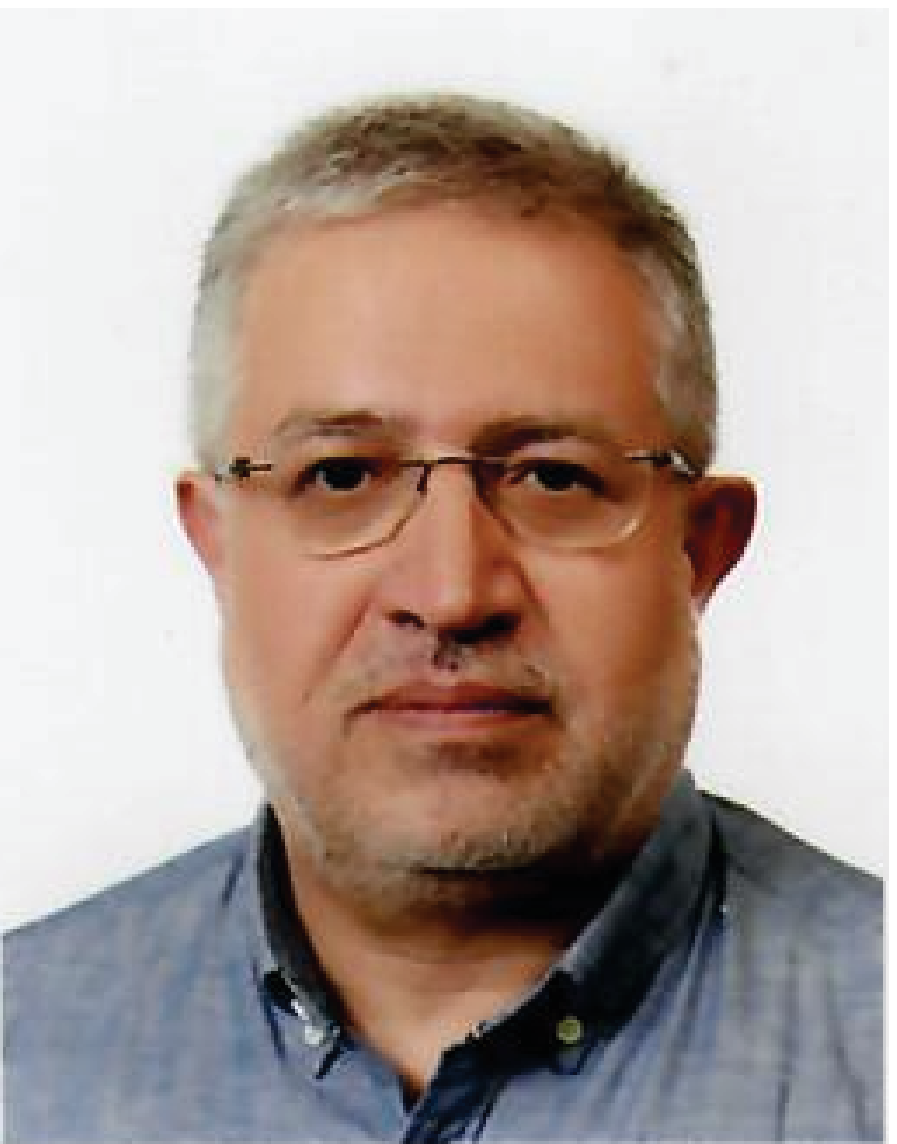}}]{Hamid R. Rabiee}
(SM'07) received his BS and MS degrees (with Great Distinction) in Electrical Engineering from CSULB, Long Beach, CA (1987, 1989), his EEE degree in Electrical and Computer Engineering from USC, Los Angeles, CA (1993), and his Ph.D. in Electrical and Computer Engineering from Purdue University, West Lafayette, IN, in 1996. From 1993 to 1996 he was a Member of Technical Staff at AT\&T Bell Laboratories. From 1996 to 1999 he worked as a Senior Software Engineer at Intel Corporation. He was also with PSU, OGI and OSU universities as an adjunct professor of Electrical and Computer Engineering from 1996-2000. Since September 2000, he has joined Sharif University of Technology, Tehran, Iran. He was also a visiting professor at the Imperial College of London for the 2017-2018 academic year. He is the founder of Sharif University Advanced Information and Communication Technology Research Institute (AICT), ICT Innovation Center, Advanced Technologies Incubator (SATI), Digital Media Laboratory (DML), Mobile Value Added Services Laboratory (VASL), Bioinformatics and Computational Biology Laboratory (BCB) and Cognitive Neuroengineering Research Center. He has also been the founder of many successful High-Tech start-up companies in the field of ICT as an entrepreneur. He is currently a Professor of Computer Engineering at Sharif University of Technology, and Director of AICT, DML, and VASL. He has been the initiator and director of many national and international level projects in the context of Iran National ICT Development Plan and UNDP International Open Source Network (IOSN). He has received numerous awards and honors for his Industrial, scientific and academic contributions. He has acted as chairman in a number of national and international conferences, and holds three patents. He is also a Member of IFIP Working Group 10.3 on Concurrent Systems, and a Senior Member of IEEE. His research interests include statistical machine learning, Bayesian statistics, data analytics and complex networks with applications in complex networks, multimedia systems, cloud and IoT privacy, bioinformatics, and brain networks.
\end{IEEEbiography}

\end{document}